\pdfoutput=1

\documentclass[11pt]{article}

\usepackage[]{acl}
\usepackage{times}
\usepackage{latexsym}
\usepackage{amsmath}
\usepackage[T1]{fontenc}

\usepackage[utf8]{inputenc}
\usepackage{microtype}
\usepackage{inconsolata}

\usepackage{tikz}
\usepackage{subcaption} 
\usepackage{amsmath}
\usepackage{multirow}
\usepackage{graphicx}
\usepackage{pgfplots}
\usepackage{adjustbox}
\usepackage{booktabs}
\usepackage{amsfonts}
\usepackage{amssymb}
\usepackage{pgfplotstable} 
\usepackage{amssymb}
\usepackage{diagbox}
\usepackage{adforn}
\usepackage{float}

\usepackage{xcolor}
\usepackage{colortbl}
\definecolor{tabhighlight}{HTML}{e5e5e5}

\newtheorem{definition}{Definition}

\title{Reconsidering Degeneration of Token Embeddings with Definitions \\ for Encoder-based Pre-trained Language Models}

\author{Ying Zhang$^{1}$, Dongyuan Li$^{2}$, and Manabu Okumura$^{1, 3}$ \\
  $^{1}$RIKEN Center for Advanced Intelligence Project \\
  $^{2}$Tokyo University \\
  $^{3}$Tokyo Institute of Technology  \\
  \texttt{ying.zhang@riken.jp \quad lidy94805@gmail.com \quad oku@lr.pi.titech.ac.jp} \\
  }

\begin{document}
\maketitle
\begin{abstract}
Learning token embeddings based on token co-occurrence statistics has proven effective for both pre-training and fine-tuning in natural language processing.
However, recent studies have pointed out that the distribution of learned embeddings degenerates into anisotropy (\textit{i.e.,} non-uniform distribution), and even pre-trained language models (PLMs) suffer from a loss of semantics-related information in embeddings for low-frequency tokens. 
This study first analyzes the fine-tuning dynamics of encoder-based PLMs and demonstrates their robustness against degeneration.
On the basis of this analysis, we propose DefinitionEMB, a method that utilizes definitions to re-construct isotropically distributed and semantics-related token embeddings for encoder-based PLMs while maintaining original robustness during fine-tuning.
Our experiments demonstrate the effectiveness of leveraging definitions from Wiktionary to re-construct such embeddings for two encoder-based PLMs: RoBERTa-base and BART-large.
Furthermore, the re-constructed embeddings for low-frequency tokens improve the performance of these models across various GLUE and four text summarization datasets.\footnote{Our code will be available at Github. 
}
\end{abstract}

\section{Introduction}
Word embeddings, served as the foundation for various NLP tasks, represent word information using low-dimensional vectors~\cite{parsing2009speech,turian-etal-2010-word}.\footnote{Please note that the embeddings discussed in this paper are the input to a model, distinct from the contextualized representations (\textit{i.e.,} the hidden states) in subsequent layers.}
To learn precise syntactic and semantic embeddings, the predominant approach is to train models based on word co-occurrence statistics~\cite{mikolov2013efficient}.
\begin{figure}[t]
    \centering
    \includegraphics[width=1\linewidth]{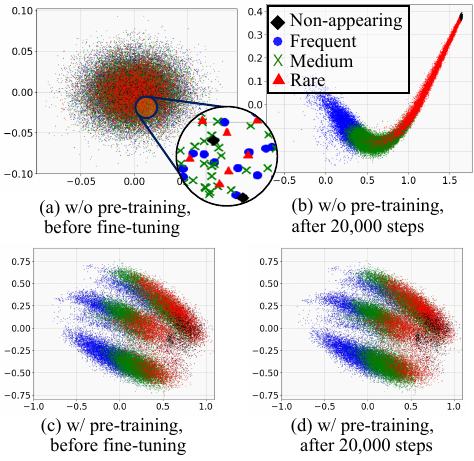}
\caption{Transformer's token embeddings with (w/) or without (w/o) pre-trained parameters during fine-tuning on the CNNDM dataset. Tokens are divided into four groups according to their frequency. (a) Isotropic distribution. (b) Narrow cone-shaped anisotropy. (c-d) BART's embeddings shape remains unchanged, indicating its robustness against degeneration into a cone.}
\label{figure:intro_isotropic}
\end{figure} 
Recent studies have highlighted that learned embeddings suffer from representation degeneration issues:  
\textit{1) Globally, the geometric distribution of the embedding matrix exhibits a narrow cone shape}~\cite{gao2018representation}. 
\textit{2) Locally, frequency-related embeddings are often mapped to the same region, making them indistinguishable and
semantically unrelated}~\cite{yu-etal-2022-rare}. 
For more easily understanding the degeneration, we used SVD to project an embedding matrix of Transformer~\citep{NIPS2017_3f5ee243} into a two-dimensional space. Figure~\ref{figure:intro_isotropic}(a) shows the distribution of the initial embeddings is isotropic (\textit{i.e.,} angularly uniform) and frequency-irrelevant. However, as shown in Figure~\ref{figure:intro_isotropic}(b), after learning task-specific embeddings, the shape changes into a narrow cone, and frequency-related embeddings are grouped together. 
The main reason 
is that the co-occurring tokens are trained to have similar representations, and other tokens are pushed in the opposite direction~\cite{tissier-etal-2017-dict2vec,mu2018allbutthetop,demeter-etal-2020-stolen}.
As a result, the expressiveness of the embedding matrix is reduced, leading to inferior performance in downstream tasks~\cite{bis-etal-2021-much}.

Previous studies have proposed solutions to alleviate the degeneration by post-processing embeddings or enhancing model optimization~\cite{mu2018allbutthetop,NEURIPS2018_e555ebe0,yu-etal-2022-rare}. 
However, these methods either handle word-level embeddings or fail to learn semantics-related embeddings for low-frequency (rare) subword tokens. This leaves the degeneration issue in token-level embeddings unsolved, particularly in pre-trained language models (PLMs)~\cite{Schick_Schütze_2020}.
Currently, PLMs with token-level vocabularies have become dominant in many NLP tasks due to their plug-and-play convenience and outstanding efficacy. 
To improve the generalization of PLMs, it is essential to explore strategies for re-constructing their embeddings.
To achieve this goal, this research focuses on encoder-based PLMs, including PLMs with encoder-only and encoder-decoder Transformer architectures, by first analyzing their behavior.
Figure~\ref{figure:intro_isotropic}(c) shows that, unlike training from scratch, the encoder-decoder PLM, BART~\citep{lewis-etal-2020-bart}, is robust against degeneration into a cone shape during fine-tuning. 
However, its pre-trained embeddings are still not ideally isotropic, with frequency-related embeddings mapped to the same region and rare tokens lacking semantics (Additional evidence is presented in $\S$~\ref{sec:token_embedding_dynamics}).

Given this, a natural question arises: \emph{how can embeddings be re-constructed for encoder-based PLMs to preserve the semantics of rare tokens, achieve isotropic distribution, and maintain robustness against degeneration during fine-tuning?}
To preserve the semantics of rare tokens, this paper proposes DefinitionEMB to re-construct their embeddings by leveraging corresponding dictionary definitions (\textit{e.g.,} Wiktionary).
Moreover, to realize effective and robust re-construction for PLMs' token-level vocabularies, DefinitionEMB uses a denoising autoencoder~\cite{10.5555/1756006.1953039} and is designed as an architecture-agnostic model that can be easily initialized from the given PLMs.
To achieve isotropic distribution across appropriate semantic regions while being robust against degeneration during fine-tuning, DefinitionEMB is trained through mimicking~\cite{pinter-etal-2017-mimicking} well-learned embeddings.
Our main contributions are summarized as follows:

\noindent $\bullet$ \textbf{Insights into Degeneration.} We observe that, although encoder-based PLMs do not degenerate into a narrow cone during fine-tuning, their pre-trained embeddings are not ideally isotropic, with frequency-related embeddings mapped to the same semantic regions.
Additionally, merely improving isotropy for encoder-based PLMs does not guarantee semantically distributed embeddings but instead can lead to further degeneration ($\S$~\ref{sec:token_embedding_dynamics}).

\noindent $\bullet$  \textbf{Embeddings against Degeneration.}
To the best of our knowledge, DefinitionEMB is the first method to re-construct pre-trained token embeddings for encoder-based PLMs being (1) isotropically distributed and semantics-related before fine-tuning, and (2) robust against degeneration into a cone shape during fine-tuning ($\S$~\ref{sec:methodology}).

\noindent $\bullet$ \textbf{Empirical Validation.}
Extensive experiments across GLUE and four text summarization datasets demonstrate the effectiveness of DefinitionEMB for RoBERTa-base and BART-large PLMs ($\S$~\ref{sec:experiments}).

\section{Related Work}
\begin{figure*}[ht]
\centering
\includegraphics[width=1\linewidth]{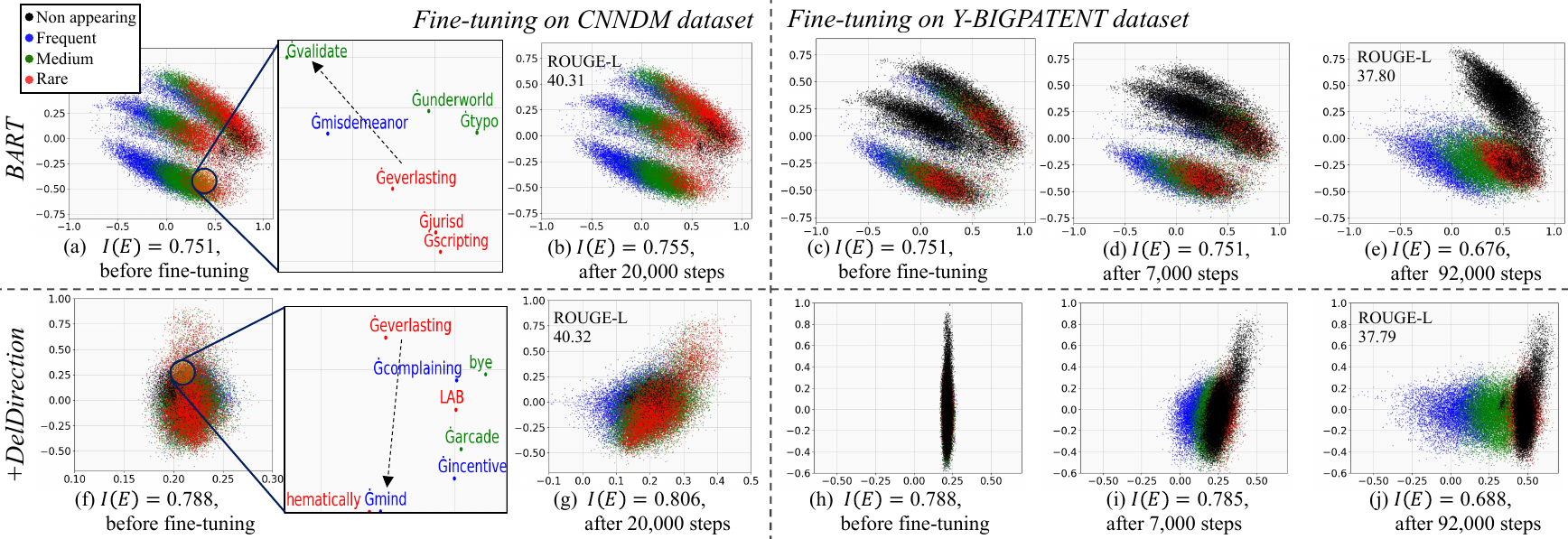}
\caption{
Token embeddings projected onto first two singular vectors.
\textbf{Top}: During fine-tuning, BART's embeddings with different frequencies maintain their distance and are robust against degeneration into a cone shape.
\textbf{Bottom}: DelDirection causes higher-frequency tokens to drift farther away from their original positions, leading to a narrow cone shape.
The dashed lines in \textbf{(a)} from ``Ġeverlasting'' point to its semantics-related tokens, recognized by both ChatGPT 3.5~\citep{achiam2023gpt} and Claude 3 Haiku~\citep{anth2024claude}. Appendix~\ref{appendix:semantically_related_llm} lists their recognition.}
\label{fig:embedding_dynamics}
\end{figure*}

\textbf{Distribution of Embeddings.} Various studies have attempted to improve the distribution of word/token embeddings. For example, 
to geometrically merge embeddings of popular (high-frequency) and rare words, \citet{NEURIPS2018_e555ebe0} utilized a discriminator of generative adversarial networks to classify embeddings as belonging to popular or rare words. 
\citet{yu-etal-2022-rare} proposed to gate the gradients of rare token embeddings to reduce degeneration into a narrow cone during training.
To achieve isotropically distributed embeddings, \citet{gao2018representation} and \citet{zhang-etal-2020-revisiting} minimized the distance between any two word embeddings.
Other studies~\cite{mu2018allbutthetop,rajaee-pilehvar-2021-cluster,bis-etal-2021-much} investigated the effectiveness of eliminating the common vector from specific groups of embeddings.
\textit{Compared to previous studies, our research considers both semantic and distributional information for embeddings.
It is also the first to apply such considerations to improve the fine-tuned performance of PLMs, whose vocabulary consists of tokenized subwords}.

\noindent \textbf{Learning Word Embeddings.} 
Learning embeddings for specific words has been discussed for a long time. Previous studies have explored various solutions, such as leveraging surface-form information~\citep{luong-etal-2013-better,pinter-etal-2017-mimicking,sasaki-etal-2019-subword} or the contexts in which these words occur~\citep{khodak-etal-2018-la,liu-etal-2019-second}.
Lexical definitions have also been considered to incorporate semantics-related information.
For instance, 
\citet{tissier-etal-2017-dict2vec} utilized definitions to bring semantically related words closer.
Various one-to-one mapping methods have also been proposed to construct embeddings for full words from corresponding definitions~\cite{bahdanau2018learning,zhang-etal-2021-language,ruzzetti-etal-2022-lacking}.
\textit{In contrast, we explore a one-to-many mapping, \textit{i.e.,} using a single definition to construct multiple token embeddings. Thus, our re-constructed embeddings can be easily applied to PLMs with token-level vocabularies.}

\noindent \textbf{Representations in PLMs.} There have also been studies that extend PLM's embeddings or analyze contextualized word representations in PLMs.
\citet{chen-etal-2022-imputing} and \citet{liang-etal-2023-graph} extended embeddings of BERT for out-of-vocabulary words. 
\citet{ethayarajh-2019-contextual} analyzed the isotropic distribution of several PLMs' contextualized word representations (\textit{i.e.,} hidden states) across model layers.
These studies are out scope of this work.

\section{Preliminaries}

Let $\mathcal{V}=\{v_{n}\}_{n=1}^{|\mathcal{V}|}$ denote the predefined restricted vocabulary of a PLM, where $|\mathcal{V}|$ tokens are ranked in \textit{descending order} according to their frequencies in pre-trained datasets. 
$\mathbf{E} \in \mathbb{R}^{|\mathcal{V}| \times h_{e}}$ denotes the pre-trained token embedding matrix,
where $h_{e}$ is the embedding size.
Given a word $\boldsymbol{w}$, we assume that $\boldsymbol{w}$ is tokenized into $K$ tokens $(v_{\boldsymbol{w},1}, \ldots, v_{\boldsymbol{w},K})$,
and $\mathbf{e}(v) \in \mathbb{R}^{h_{e}}$ denotes the embedding of token $v$.

\textcolor{black}{The geometry of $\mathbf{E}$ is assumed to capture linguistic regularities, \textit{i.e.,} the similarity between token embeddings reflects the semantic similarity of the corresponding tokens.
Therefore, researchers expect $\mathbf{E}$ to exhibit a uniform distribution, denoted as isotropic, to maximize the containment of linguistic information for distinguishing tokens.
To estimate isotropy,
\citet{mu2018allbutthetop} proposed the metric $I(\mathbf{E})$$\in$$[0,1]$, where a value closer to 1 indicates higher isotropy. 
They also proposed a post-processing technique to improve isotropy for $\mathbf{E}$ by eliminating the common mean vector and top-$\beta$ dominating directions from $\mathbf{E}$.
\textit{We denote this method as \textbf{DelDirection} and consider it as a baseline with $\beta$=10.} Appendix~\ref{appendix:isotropy_metric} shows more details.}

\section{Token Embedding Dynamics: An Experimental Investigation}
\label{sec:token_embedding_dynamics}

Inspired by~\citet{yu-etal-2022-rare}'s observation that Transformer exhibits narrow cone-shaped degeneration due to the unbalanced token frequency, we investigated whether token frequency influences the embedding of PLMs during fine-tuning. 
We first classified tokens into appearing and non-appearing groups based on their appearance in the corresponding fine-tuning dataset.
Then, 30\%, 50\%, and 20\% appearing tokens in $\mathcal{V}$ were assigned to the frequent, medium, and rare groups, respectively.
The main paper primarily focuses on the BART-large model on the CNNDM and Y-BIGPATENT datasets, where CNNDM has a large unique appearing vocabulary, while Y-BIGPATENT has a small one.
Appendices~\ref{appendix:vocab_distribution} and \ref{appendix:projected_embeddings} show more details.

In Figure~\ref{fig:embedding_dynamics}, we conclude that \textbf{the encoder-based PLM does not degenerate into a narrow cone; instead, it exhibits drift patterns influenced by token frequency.}
\textbf{Additionally, merely improving isotropy for it does not guarantee semantically distributed embeddings and does not improve performance for downstream tasks.}
We would like to share the following three main findings:

\vspace{4pt}
\noindent \textbf{Finding-1}. 
\textit{The pre-trained embeddings of BART still suffer from anisotropy and a lack of semantics.} Before fine-tuning, BART achieves an $I(\mathbf{E})$ score of 0.751, suggesting room for improvement. The zoomed-in view in Figure~\ref{fig:embedding_dynamics} (a) further shows that rare tokens still lack sufficient semantic information, {\textit{e.g.,} most neighbors of the rare token ``Ġeverlasting'' are semantically unrelated to it. This finding motivates us to incorporate semantics-related information and improve isotropy before fine-tuning to enhance downstream performance.

\vspace{4pt}
\noindent \textbf{Finding-2}. \textit{Although using DelDirection before fine-tuning yields higher $I(\mathbf{E})$ and more thoroughly mixed embeddings, it leads to a fragile distribution and no further improved ROUGE~\citep{lin-2004-rouge} scores on the downstream datasets.}
Specifically, as seen in Figures~\ref{fig:embedding_dynamics} (g) and (i), popular tokens drift away from the original overlapping position. 
After more updates, as seen in Figure~\ref{fig:embedding_dynamics} (j), tokens degenerate into a narrow cone with frequency bias.
Finally, it does not improve model performance (\textit{i.e.,} ROUGE). This motivates us to investigate an effective method to alleviate degeneration during fine-tuning and improve performance.

\vspace{4pt}
\noindent \textbf{Finding-3}. \textit{Figures~\ref{fig:embedding_dynamics} (b), (d), and (e) show that, while appearing and non-appearing tokens drift into different regions, BART is robust against degeneration into a cone shape with increasing number of updates.}
This robustness is neither architecture- nor size-dependent, as RoBERTa-base exhibits similar robustness against degeneration (see Appendix~\ref{appendix:projected_embeddings}).
This finding aligns with recent studies which suggest that fine-tuning rarely alters the majority of PLMs' parameters~\cite{jain2024mechanistically,pmlr-v202-panigrahi23a}. These results motivate us to leverage the robustness of PLMs to prevent degeneration into a cone shape during fine-tuning.

\section{Methodology}
\label{sec:methodology}
\begin{figure*}[t]
    \centering
    \includegraphics[width=1\linewidth]{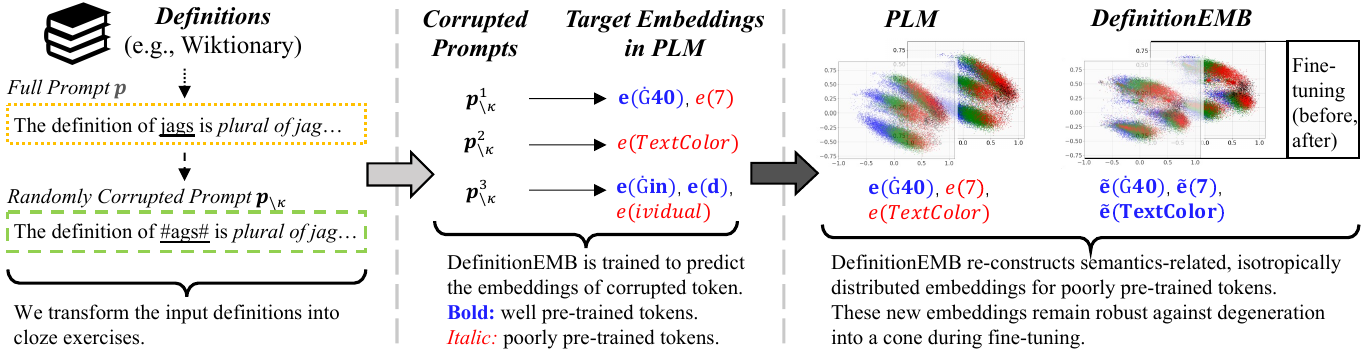}
    \caption{Overview of constructing definition embeddings to replace pre-trained embeddings.}
    \label{fig:overview_of_procedure}
\end{figure*}

Let $f(\theta^{\text{PT}})$ be an encoder-based PLM with parameter $\theta^{\text{PT}}$. As discussed in $\S~\ref{sec:token_embedding_dynamics}$, the pre-trained token embeddings $\mathbf{E}$ of $f(\theta^{\text{PT}})$ suffer from anisotropy and a lack of semantics before fine-tuning. The re-constructed embeddings should also ensure robustness against degeneration during fine-tuning.
As shown in Figure~\ref{fig:overview_of_procedure}, we introduce a novel framework \textbf{DefinitionEMB} to solve these issues by simply re-constructing token embeddings for $\mathbf{E}$ from the corresponding word definition.

\subsection{Semantic Embedding Re-construction}

Recall that pre-trained embeddings of rare token still lack semantic information. \citet{ausubel1978educational} indicated that achieving a deep understanding of a new word involves linking it to already known relevant concepts. Motivated by this, DefinitionEMB connects tokens of a word to their corresponding dictionary definitions (\textit{e.g.,} Wiktionary).
To help DefinitionEMB effectively understand semantics of rare tokens, we base our approach on a denoising autoencoder, which first corrupts a sequence into a noisy version and then re-constructs it back to a clean form~\cite{10.5555/1756006.1953039}.
Thus, embedding re-construction is realized as a cloze exercise, as shown in the left of Figure~\ref{fig:overview_of_procedure}.

To form the cloze task, we first map a word $\boldsymbol{w}$ and its definition into a sentence, referred to as the full prompt $\boldsymbol{p}$.
Then, we randomly corrupt parts of tokens in the word $\boldsymbol{w}$ within the full prompt and train DefinitionEMB to re-construct the embeddings of these corrupted tokens.
Note that in this example, ``\#'' could be filled with multiple subwords tokens. Several tokens, such as ``ags'', are randomly provided in the underline, with the underlined tokens ``\#ags\#'' corresponding to the target word ``jags''. Since the embedding of the random token ``ags'' may be noisy, this denoising strategy helps to generate effective representations when facing noisy inputs.
In addition, DefinitionEMB is designed as an architecture-agnostic model, that can be easily initialized from the given PLMs to leverage pre-trained parameters.
We apply different corruption strategies based on model types: BERT-style masking~\cite{devlin-etal-2019-bert} for encoder-only PLMs and T5-style masking~\cite{10.5555/3455716.3455856} for encoder-decoder PLMs.
Their effectiveness has been well demonstrated in prior studies~\cite{voita-etal-2019-bottom,10.5555/3455716.3455856,salazar-etal-2020-masked}.

Mathematically, let $\kappa$ denote the set of corrupted positions, $\boldsymbol{p}_{\kappa} \subseteq \boldsymbol{w}$ the set of corrupted tokens, and $\boldsymbol{p}_{\backslash \kappa}$ the corrupted prompt.
Let $v_{\boldsymbol{p},i}$ represent the $i^{\text{th}}$ token in prompt $\boldsymbol{p}$.
For a token $v_{\boldsymbol{p},k}$, where $k \in \kappa$, 
DefinitionEMB re-constructs its semantics-related definition embedding $\tilde{\mathbf{e}}(v_{\boldsymbol{p},k})$ as follows:
\vspace{-7pt}
\begin{align}
    \mathbf{s}_{g(k)} &=f(k,  \boldsymbol{p}_{\backslash 
 \kappa};\theta^{\text{PT}}), \nonumber \\ 
    \tilde{\mathbf{e}}(v_{\boldsymbol{p},k}) &= \mathbf{W} \mathbf{s}_{g(k)}, \label{eq:reconstruction_embedding}
\end{align}
where $\mathbf{s}_{g(k)} \in \mathbb{R}^{h_{s}}$ is the last hidden state at position $g(k)$ and $\mathbf{W} \in \mathbb{R}^{h_{e} \times h_{s}}$ is a weight matrix.
The function $g(\cdot)$ maps the corrupted position to the prediction position. For an encoder-only PLM, $g(k) = k$. For an encoder-decoder PLM, $g(k)$ refers to the position of the delimiter in front of $v_{k}$ in the decoder.
Appendix~\ref{appendix:corrupted_prompts} explains our methods for generating and corrupting prompts, and $g(\cdot)$.

\subsection{Robustness against Degeneration}
Definitions tend to contain popular and easily understandable words to explain their corresponding target words. Thus, the pre-trained embeddings for most tokens in the cloze exercise can be isotropically distributed and semantics-related. 
DefinitionEMB is, therefore, trained using a mimicking approach~\cite{pinter-etal-2017-mimicking}, which optimizes embeddings by matching the predicted embeddings to well pre-trained ones.
Through finding the overall optimal embedding space for the PLM's vocabulary $\mathcal{V}$, DefinitionEMB learns a transformation function from ideally distributed input embeddings to target embeddings.
This approach ensures that the re-constructed embeddings of poorly pre-trained tokens are uniformly distributed across appropriate semantic regions while being robust against degeneration during fine-tuning.
Next, we discuss how to identify well pre-trained embeddings. 
Someone might consider directly adopting embeddings of high-frequency tokens. However, we argue that embeddings of rare tokens may also contain useful information.
Therefore, following~\citet{burns2023discovering}, we incorporate definitions involving all tokens in $\mathcal{V}$ for training, allowing the model to automatically identify well pre-trained embeddings.

Let $\mathcal{D}=\{(\boldsymbol{w}^{m}, \boldsymbol{p}^{m})\}^{M}_{m=1}$ denote a corpus with $M$ word-definition pairs. 
For $v_{\boldsymbol{p}^{m},k}$, DefinitionEMB is optimized by minimizing the mean squared error between pre-trained embeddings $\mathbf{e}(v_{\boldsymbol{p}^{m},k})$ and definition embeddings $\tilde{\mathbf{e}}(v_{\boldsymbol{p}^{m},k})$ as:
\vspace{-7pt}
\begin{equation}
    \hspace*{-4pt} 
    \mathcal{L} = \sum_{m=1}^{M} \frac{\sum_{k \in \kappa^{m}}\left \|\mathbf{e}(v_{\boldsymbol{p}^{m},k}) - \tilde{\mathbf{e}}(v_{\boldsymbol{p}^{m},k})\right \|^{2}}{M|\kappa^{m}|}, \label{eq:objective_function}
\vspace{-7pt}
\end{equation}
where $|\kappa^{m}|$ is the number of corrupt tokens for $\boldsymbol{p}^{m}$.

\subsection{Replacing Strategy in Inference}
\label{sec:replacing_strategy}

Given a specific downstream task, we first remove tokens from $\mathcal{V}$ that do not appear in the corresponding fine-tuning dataset. The remaining set of tokens is denoted as $\mathcal{V}_{[task]}$, where the tokens in $\mathcal{V}_{[task]}$ are ranked in \textit{descending order} according to their frequency in the dataset for pre-training. Our preliminary experiments in Appendix~\ref{appendix:pre_experiment} demonstrate that, when replacing $5\%$ of the last tokens in $\mathcal{V}_{[MRPC]}$, BART achieves the highest accuracy compared to replacing random or top tokens in $\mathcal{V}_{[MRPC]}$ on MRPC dataset.
Therefore, as an initial study on re-constructing token embeddings for PLMs, this study focuses on low-frequency tokens. 
Our replacing strategy has the following definition. 
\begin{definition}\label{definition-1}
For any PLMs with diverse data distributions and task requirements for downstream tasks, we replace $\min(\alpha \% * |\mathcal{V}|, |\mathcal{V}_{[task]}|)$ of the last tokens in $\mathcal{V}_{[task]}$, where $\alpha$ 
is a hyperparameter.
\end{definition}

Based on Definition~\ref{definition-1}, \textcolor{black}{before fine-tuning,} we straightforwardly replace $\min(\alpha \% * |\mathcal{V}|, |\mathcal{V}_{[task]}|)$ pre-trained embeddings with definition embeddings as $\mathbf{e}(v) = \tilde{\mathbf{e}}(v)$, which can avoid missing definition information or excessive noise in $\mathbf{e}(v)$.

\section{Experiments}
\label{sec:experiments}
Following~\citet{mu2018allbutthetop} and~\citet{lewis-etal-2020-bart}, we assessed the performance of PLMs with diverse embeddings on various benchmark tasks, including embeddings isotropy, word similarity, and natural language understanding on the General Language Understanding Evaluation (GLUE) benchmark. Additionally, for the encoder-decoder PLM, we tested its performance with replaced embeddings on the text summarization task to evaluate both its text comprehension and generation capabilities.
Specifically, GLEU and text summarization evaluated the fine-tuned performance of PLMs.

\begin{table}[t]
\centering
\resizebox{1\linewidth}{!}{
\begin{tabular}{lcccc}
\toprule
\multirow{2}{*}{\textbf{Model}} & \multicolumn{4}{c}{\textcolor{black}{$I(\mathbf{E})$ $\uparrow$}}\\
\cline{2-5}
&\textbf{Frequent}  &\textbf{Medium} &\textbf{Rare}  &\textbf{All Tokens} \\
\midrule
\midrule
RoBERTa & \textbf{0.694} & 0.501 & 0.315 & 0.504 \\
+ DelDirection & 0.639 & \textbf{0.641} & \textbf{0.599} & \textbf{0.624} \\
+ DefinitionEMB & 0.649 & 0.470 & 0.382 & 0.519 \\
\midrule
BART                   &  \textbf{0.851}     &   0.668      &  0.515    &  0.751   \\
+ DelDirection  &   0.790    &     0.775    &  \textbf{0.731}    
&  0.788  \\
\rowcolor{tabhighlight} + DefinitionEMB & 0.834 & \textbf{0.800} & 0.603 & \textbf{0.876} \\
\bottomrule
\end{tabular}
}
\caption{Isotropy of $\mathbf{E}$. The frequent (30\%), medium (50\%), and rare (20\%) groups are determined based on the token index in $\mathcal{V}$.
Appendix~\ref{appendix:projected_initial_token_embeddings} shows projected $\mathbf{E}$.}
\label{table:isotropy_results}
\end{table}

\subsection{Experimental Settings}

For different downstream tasks, we used various datasets, as described in the following experimental sections. The statistics for these datasets are described in Appendix~\ref{appendix:vocab_distribution}.
We adopted token-level DelDirection~\cite{mu2018allbutthetop} as our baseline since all other post-processing methods focused on word-level degeneration issues and could not be applied to PLMs.
We adopted the RoBERTa-base and BART-large models as our backbone PLMs with encoder-only and encoder-decoder architectures, respectively. Both backbones used the same vocabulary $\mathcal{V}$ with size $N=50,265$. 
Appendix~\ref{appendix:hyperparameters} lists the hyperparameter settings, tuning\footnote{We tuned $\alpha$ on the validation set.} for DefinitionEMB, and fine-tuning PLMs for downstream tasks.
Scores are the average over three trials.

We used 1.5G English-language Wiktionary as definitions to train DefinitionEMB. The 1,464,327 extracted definitions were randomly divided into 1,454,327 for training and 10,000 for validation.
We also extracted 1,388 definitions from the Internet used for numbers and named entity tokens.
Overall, 1,455,715 examples were used for training DefinitionEMB. 
During inference, these examples were reused for loading definition embeddings. Once a token embedding has been replaced, it will not be replaced during the rest of the procedure. 
2,305 tokens from $\mathcal{V}$ were always excluded from replacement because they did not have a corresponding definition, such as ``)=(''.

\subsection{Quantitative Evaluation}
\noindent \adfhalfrightarrowhead \, \textbf{Initial Isotropy.} We measured the initial isotropy of token embeddings $\mathbf{E}$ in \textcolor{black}{PLMs} and the isotropy after completely replacing $\mathbf{E}$. 
Table~\ref{table:isotropy_results} shows that both RoBERTa and BART exhibit high isotropy in the frequent group but low isotropy in the rare group. DelDirection helps achieve a uniformly distributed isotropy across frequency groups and results in the highest isotropy for the rare group, although it decreases isotropy in the frequent group.
DefinitionEMB also showcases a more uniform isotropy distribution than the original PLMs, displaying a lower isotropy for the frequent group but higher isotropy for the rare group.
Additionally, DefinitionEMB for BART achieves the highest isotropy for the medium group as well as for the entire $\mathbf{E}$.
Appendix~\ref{appendix:mse_between_embeddings} analyzes differences of using DefinitionEMB on BART and RoBERTa due to the different distribution of pre-trained embeddings and used masking strategies.

\begin{table}[t]
\centering
\resizebox{1\columnwidth}{!}{
\begin{tabular}{lccccc}
\toprule
\multirow{2}{*}{\textbf{Model}} & \multicolumn{5}{c}{\textcolor{black}{\textbf{Spearman Score} $\uparrow$}}\\
\cline{2-6}
& \textbf{RG65}  & \textbf{RW}   & \textbf{SimLex}  & \textbf{SimVerb}  & \textbf{Ave}\\
\midrule
\midrule
RoBERTa & 16.05 & 18.89 & 26.67 & 11.81 & 18.36 \\
+ DefinitionEMB & \textbf{18.88} & \textbf{18.96} & \textbf{27.15} & \textbf{11.91} & \textbf{19.23}\\
\midrule
BART       & 15.32    & 19.66     &      28.56      & \textbf{13.09} & 19.16\\
\rowcolor{tabhighlight} + DefinitionEMB & \textbf{15.67} & \textbf{19.76}   &      \textbf{28.63}            & 12.72 & \textbf{19.20}\\
\bottomrule
\end{tabular}
}
\caption{Results on the word similarity task with dot product.
DefinitionEMB completely replaced $\mathbf{E}$.
Appendix~\ref{appendix:word_similarity} shows the results with cosine similarity.}
\label{table:word_similarity}
\end{table}

\begin{table*}[t]
\centering
\footnotesize
\setlength{\tabcolsep}{3.2mm}{
\begin{tabular}{lccccccccc}
\toprule
\multirow{2}{*}{\textbf{Model}} & \multirow{2}{*}{\textbf{SST}} & \multirow{2}{*}{\textbf{MRPC}} & \multirow{2}{*}{\textbf{STS}} & \multirow{2}{*}{\textbf{QQP}} & \multicolumn{2}{c}{\textbf{MNLI}} & \multirow{2}{*}{\textbf{QNLI}} & \multirow{2}{*}{\textbf{RTE}} & \multirow{2}{*}{\textbf{Average}} \\
&                               &                                &                                 &                               & \textbf{m}      & \textbf{mm}     &                                &      &                        \\
\midrule
\midrule
RoBERTa & 95.7 & 87.5 & \textbf{89.6} / \textbf{89.0} & \textbf{89.6} & 87.3 & 86.7& \textbf{93.1} & 73.9 & 88.0\\
+ DelDefinition & \textbf{95.9} & 86.9 & 89.0 / 88.3 & 89.3 & 87.3 & 86.8 & 93.0 & 72.3 & 87.6\\
+ DefinitionEMB & \textbf{95.9} & \textbf{87.7} & \textbf{89.6} / \textbf{89.0} & 89.4 & \textbf{87.6} & \textbf{87.0} & 93.0 & \textbf{75.3} & \textbf{88.3}\\
\midrule
BART                            & \textbf{96.5}                          & 87.8                           & 91.2 / 90.6                       & \textbf{90.1}                          & \textbf{90.0}            & \textbf{89.2}            & 94.7                           & 82.4                         &  90.3\\
+ DelDefinition & 96.4 & 87.3 & 90.9 / 90.4 & 89.9 & 89.9 & \textbf{89.2} & 94.7 & 78.7 & 89.7\\
\rowcolor{tabhighlight} + DefinitionEMB                                & 96.4                          & \textbf{88.3}                           & \textbf{91.3} / \textbf{90.7}                       & \textbf{90.1}                              & \textbf{90.0}            & \textbf{89.2}            & \textbf{94.9}                           & \textbf{83.3}  &  \textbf{90.5}\\
\bottomrule
\end{tabular}
}
\caption{Experimental results on GLUE. For the STS dataset, we report the Pearson/Spearman's rank correlation, while for other datasets, we report accuracy scores. For the MNLI dataset, we report results for Matched (m) and Mismatched (mm) sets.
Appendix~\ref{appendix:isotropy_on_GLUE} describes corresponding $I(\mathbf{E})$.
}
\label{table:GLUE_results}
\end{table*}

\begin{table*}[t]
\centering
\resizebox{1\linewidth}{!}{
\begin{tabular}{lccccc}
\toprule
\multirow{2}{*}{\textbf{Model}}     & \multirow{2}{*}{\textbf{CNNDM}} & \multirow{2}{*}{\textbf{Y-BIGPATENT}} & \multirow{2}{*}{\textbf{XSum}} & \multirow{2}{*}{\textbf{Billsum}} & \multirow{2}{*}{\textbf{Average}} \\
& & & & &\\
\midrule
\midrule
BART               &  43.57 \, / 20.93 / 40.31 \,  & 43.96 \, / 18.92 \, / 37.80 \,  & 43.76 \, / 20.40 \, / 34.65 \, & \textbf{51.02} / 32.44 \, / 39.11 & 45.58 / 23.17 / 37.97\\
+DelDirection  & 43.59 \, / 20.93 / 40.32 \,  & 43.91 \, / 18.85 \, / 37.79 \,  & 43.90 \, / 20.58 \, / 34.86 \,  & 50.89 / 32.22 \, / 38.97  &   45.57 / 23.15 / 37.99      \\
\rowcolor{tabhighlight} +DefinitionEMB & \textbf{43.78}$\dagger$ / \textbf{20.94} / \textbf{40.52}$\dagger$ &  \textbf{44.16}$\ddagger$ / \textbf{19.06}$\ddagger$ / \textbf{38.01}$\ddagger$  &  \textbf{43.96}$\dagger$ / \textbf{20.61}$\dagger$ / \textbf{34.87}$\dagger$  & 50.96 / \textbf{32.64}$\ddagger$ / \textbf{39.28} & \textbf{45.72} / \textbf{23.31} / \textbf{38.17} \\
\bottomrule
\end{tabular}
}
\caption{Experimental results (ROUGE1-F1 / ROUGE2-F1 / ROUGEL-F1) on the text summarization tasks.
$\dagger$ and $\ddagger$ indicate that the score is significantly superior to BART with a p-value < 0.01 and < 0.05, respectively.
}
\label{table:text_summarization_results}
\end{table*}

\begin{figure*}[t]
    \centering
    \includegraphics[width=1\linewidth]{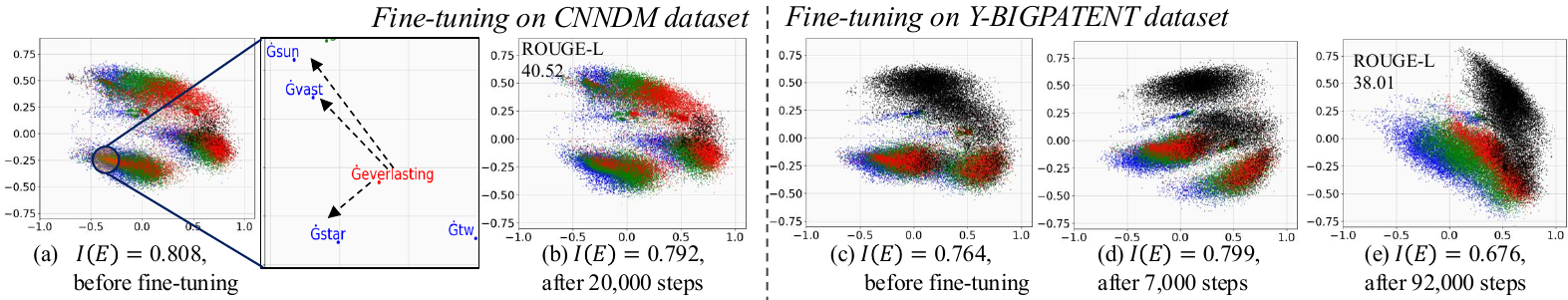}
\caption{Projected token embeddings in BART+DefinitionEMB before and after fine-tuning. 
Embeddings with different frequencies maintain their distance during fine-tuning, demonstrating robustness against degeneration.
The embeddings in (a) and (c) exhibit different shapes due to the different $\alpha$.}
\label{fig:embedding_dynamics_definitionEMB}
\end{figure*}

\noindent \adfhalfrightarrowhead \, \textbf{Word Similarity.} Compared to BART and RoBERTa, which utilize over 160 GB of contexts to learn semantic embeddings, DefinitionEMB utilizes definition information at only 1\% of the size of their contexts.
To evaluate the construction capabilities of DefinitionEMB on representations given such limited information, we adopted the word similarity task to investigate whether DefinitionEMB can maintain the original semantic relationships.\footnote{DelDirection was excluded because it does not construct representations but rather updates existing embeddings.}
Following~\citet{mu2018allbutthetop}, we assessed whether the similarity between the embeddings of two given words aligns with the ground truth, in terms of Spearman’s rank correlation.
We used the dot product to measure the similarity between word embeddings across four datasets: RG65, rare-words (RW), SimLex-999, and SimVerb-3500.
We estimated the word embeddings by summing the embeddings of the corresponding tokens.
As Table~\ref{table:word_similarity} shows, using DefinitionEMB yields higher Spearman scores than the original PLMs on the RG65, SimLex-999, and RW datasets.
In particular, the results of the RW dataset, which consists of only rare words, underscore the effectiveness of DefinitionEMB in capturing the semantic information for these words.

\noindent \adfhalfrightarrowhead \,\textbf{GLUE.} \: 
We conducted experiments on the GLUE benchmark~\citep{wang-etal-2018-glue} across seven datasets: 
SST, MRPC, STS, QQP, MNLI, QNLI, and RTE.
Table~\ref{table:GLUE_results} reports our test set results obtained from the public leaderboard.\footnote{\url{https://gluebenchmark.com/}}
Using DelDirection results in the lowest accuracy and Pearson/Spearman’s rank correlation in most cases; however, it achieves the highest $I(\mathbf{E})$ across all datasets, as listed in Appendix~\ref{appendix:isotropy_on_GLUE}.
This supports our assumption that DelDirection focuses on the global isotropic distribution of embeddings at the expense of semantic information.
In contrast, DefinitionEMB improves both the global distribution and downstream performance on GLUE, particularly for the MRPC and RTE datasets. This improvement highlights DefinitionEMB's advantage in enhancing semantic information, as these tasks contain low-frequency tokens that suffer from insufficient fine-tuning, thus increasing the necessity for replacing their embeddings.\footnote{While the STS task also involves low-frequency tokens, it primarily requires an understanding of popular words, such as the distinction between ``woman'' and ``man''. In contrast, MRPC and RTE involve the comprehension of more complex sentences. This difference explains why DefinitionEMB does not show the same level of improvement across these tasks.}

\noindent \adfhalfrightarrowhead \, \textbf{Text Summarization.} \: For the downstream summarization task, we used public abstractive summarization datasets, including CNN/DailyMail (CNNDM), Extreme Summarization (XSum), BillSum, and Y-BIGPATENT.
We evaluated the model performance on these datasets using the ROUGE scores and compared BART + DefinitionEMB with the original BART using paired bootstrap resampling~\citep{koehn2004statistical} for the significance test.
Table~\ref{table:text_summarization_results} shows that using DelDirection improves the ROUGEL-F1 score by 0.21 points for BART on the XSum dataset. However, the difference between BART and DelDirection is very limited in other datasets; DelDirection achieves even a lower ROUGEL-F1 score, with a decrease of 0.14 points, than BART on the Billsum dataset.
In contrast, by improving semantics-related information and frequency-aware $I(\mathbf{E})$ for rare tokens, DefinitionEMB achieves the highest ROUGEL-F1 scores, 
with improvements of 0.21, 0.21, 0.22, and 0.17 points on the CNNDM, Y-BIGPATENT, XSum, and Billsum datasets, respectively, for BART.

\subsection{Analysis of DefinitionEMB}

\begin{table}[t]
\resizebox{1\columnwidth}{!}{
\begin{tabular}{llll}
\toprule
\multirow{2}{*}{\textbf{Models}} & \multicolumn{3}{c}{\textbf{ROUGE (F1)} $\uparrow$}   \\
& ROUGE-1 & ROUGE-2 & ROUGE-L\\
\midrule
\midrule
BART            & 34.99              & 14.67              & 32.19              \\
+DelDirection   & 35.13 (+0.14)              & 15.05 (+0.38)              & 32.33 (+0.14)              \\
\rowcolor{tabhighlight} +DefinitionEMB  & \textbf{36.11$\ddagger$(+1.12)}              & \textbf{15.75$\dagger$(+1.08)}              & \textbf{33.44$\ddagger$(+1.25)}              \\ \bottomrule
\end{tabular}
}
\caption{\textcolor{black}{Experimental results on the CNNDM subset for rare tokens (index > 40,000 in $\mathcal{V}$). (+scores) indicates the improvement compared to BART.}}
\label{table:filtered_cnndm}
\end{table}

\noindent \adfhalfrightarrowhead \, \textbf{Ablation Study.} Appendix~\ref{appendix:ablation_mrpc} analyzes the effectiveness of replacing only appearing tokens instead of all tokens.
To evaluate DefinitionEMB's ability to handle rare tokens, we conducted experiments on 65 data pairs of the CNNDM test set, whose target sentence consisted of a high proportion of rare tokens (at least 5\%). Details of this subset are provided in Appendix~\ref{appendix:hyperparameters}. As shown in Table~\ref{table:filtered_cnndm}, DefinitionEMB achieves superior scores, with improvements of over 1 point across all ROUGE metrics, compared to BART, while DelDirection improves only ROUGE1 and ROUGEL by 0.14 points. These results highlight the effectiveness of DefinitionEMB and show that DelDirection, as an update method rather than a construction method for embeddings, is less effective for rare tokens on downstream tasks.

\begin{figure}[t]
    \centering
    \includegraphics[width=1\linewidth]{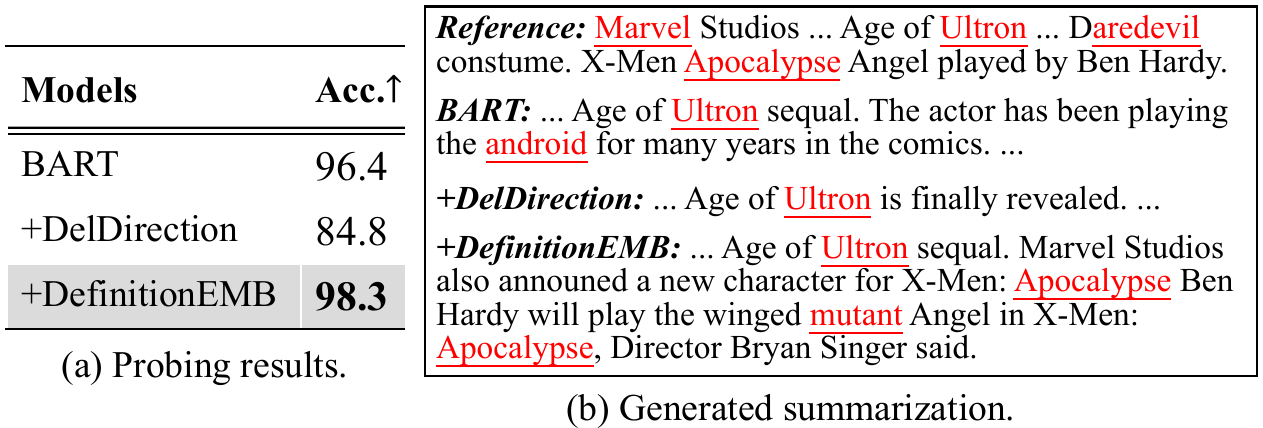}
\caption{
Case study on the CNNDM dataset. 
(a) Probing results.
(b) Example generated summaries.
\textcolor{red}{\underline{Underline}} indicates the rare tokens with index larger than 40,000 in $\mathcal{V}$.  
Appendix~\ref{appendix:sample_of_summarization} includes a full version.}
\label{figure:probing_results}
\end{figure}

\noindent \adfhalfrightarrowhead \, \textbf{Embedding Dynamics.} Figure~\ref{fig:embedding_dynamics_definitionEMB} depicts the projected token embeddings of DefinitionEMB.
On the CNNDM dataset, tokens exhibit minimal drift from ``before'' to ``after'' fine-tuning. Conversely, on the Y-BIGPATENT dataset, tokens within the same group move together after fine-tuning. 
These findings align with those of BART in Figure~\ref{fig:embedding_dynamics}, indicating that using DefinitionEMB helps maintain BART's robustness against degeneration into a narrow cone. Additionally, using DefinitionEMB before fine-tuning increases $I(\mathbf{E})$ for BART across the CNNDM and Y-BIGPATENT datasets. It supports our observation that, compared to BART, embeddings in BART+DefinitionEMB with different frequencies are more thoroughly mixed.

\noindent \adfhalfrightarrowhead \, \textbf{Probing Semantics.}  Zooming in Figure~\ref{fig:embedding_dynamics_definitionEMB} (a) reveals that the token ``Ġeverlasting'' is surrounded by more semantically related tokens than BART with and without DelDirection before fine-tuning (Figure~\ref{fig:embedding_dynamics}).
This case study demonstrates that using DefinitionEMB provides more semantically related information in the initial step (\textit{i.e.,} the input layer). To investigate whether a model can encode and extract such semantic information in subsequent layers,
we followed~\citet{allen-zhu2024physics} and conducted a simple probing test.
We freezed each CNNDM fine-tuned model and trained a classifier to classify the last layer’s hidden states of input words as ``numeric < 1,000'', ``numeric > 1,000'', or ``others''. In inference, we evaluated accuracy on 78 rare numeric tokens and 78 rare non-numeric tokens. Higher accuracy indicates a stronger link to the token’s semantics. The overall accuracy is shown in Figure~\ref{figure:probing_results} (a). Using BART+DefinitionEMB achieves the highest accuracy at 98.3\%, demonstrating that it encodes and extracts more semantic information. Appendix~\ref{appendix:details_on_probing} shows more details.

\section{Conclusion}
This research found that during fine-tuning, the embeddings of the encoder-based PLMs do not degenerate into a narrow cone in a low-dimensional space.
However, eliminating specific directions from embeddings to ensure isotropy leads to loss of semantics and further degeneration.
Experimental results demonstrated that using DefinitionEMB for 
PLMs, RoBERTa and BART, improves the distribution of embeddings and enables low-frequency token embeddings to retain semantics-related information before fine-tuning.
These new embeddings also maintained PLMs' robustness against degeneration and led to improved downstream performance.
Considering the recent emergence of understanding the internal workings of language models, this research paves the way for future work to explore why PLMs are robust against degeneration.

\section*{Limitations}
\textbf{Limited Models.} The scope of this paper is limited to encoder-based PLMs. Although DefinitionEMB can yield effective embeddings for these models by the reconstruction, its denoising autoencoder-based objective function makes it challenging to directly apply it to decoder-only PLMs. This is because decoder-only PLMs use standard causal language modeling (left-to-right), which causes different information flow across layers compared to the masked language modeling, utilized in DefinitionEMB~\cite{voita-etal-2019-bottom}. While we aimed to emphasize the effectiveness of definition datasets in addressing degeneration for PLMs, future work could explore methods that can be effectively applied to decoder-only PLMs, especially large-scale ones.\newline
\newline
\noindent \textbf{Limited Understanding of PLMs' Robustness.} One of our findings is that encoder-based PLMs do not degenerate into a cone shape during fine-tuning. While this finding aligns with recent studies that PLMs rarely alter their parameters during fine-tuning~\cite{jain2024mechanistically,pmlr-v202-panigrahi23a}, it tells only half the story — modifying model parameters before fine-tuning can lead to more effective performance. However, to fully understand why PLMs' embeddings are robust against degeneration during fine-tuning and to pinpoint improvements in the models, the interplay between embedding layers and other components of the model architecture, such as attention layers, could be further studied.

\section*{Acknowledgements}
We thank Prof. Min-Yen Kan for suggesting the types of rare tokens to us. We thank Prof. Keisuke Sakaguchi and Prof. Sho Yokoi for their suggestions on the writing.

\bibliography{anthology,custom}

\appendix

\clearpage

\pgfplotsset{width=7cm,compat=1.15}
\pgfplotstableread[row sep=\\,col sep=&]{
    Position & MRPC & CNNDM & Billsum & BIGPATENT & XSum \\
    1 & 4449 & 4995 & 4883 & 3830 & 4992 \\
    2 & 3061 &   4998 &   4673 &   3305 & 4992 \\
    3 & 2041 &   4996 &  4416   & 3081 & 4980 \\
    4 & 1454 &   4995 &   4200  &  3047 & 4981\\
    5 & 1149 &   4995 &   3967  &  3080 & 4968\\
    6 & 902 & 4997 &   3895 &   3163 & 4959 \\
    7 & 660 & 4994 &   3821 &   3193 & 4942\\
    8 & 561 & 4985 &   3617 &   3045 & 4912 \\
    9 & 367 & 4963 &   3169 &   2694 & 4825 \\
    10 & 119 & 4065 &   1582 &   2160 & 3318 \\
    }\MainVocabDistribution

\pgfplotstableread[row sep=\\,col sep=&]{
    Position & MRPC & CNNDM & Billsum & BIGPATENT & XSum \\      
    1 & 1.535 &  4.647 &  4.073 &  4.964 & 4.231 \\
    2 & 0.613 &  3.566 &  2.757 &  4.024 & 3.141\\
    3 & 0.477 &  3.276 &  2.425 &  3.859 & 2.831 \\
    4 & 0.415 &  3.076 &  2.279 &  3.622 & 2.623 \\
    5 & 0.380 &  2.914 &  2.214 &  3.531 & 2.466 \\
    6 & 0.380 &  2.772 &  2.055 &  3.644 & 2.329 \\
    7 & 0.342 &  2.612 &  1.967 &  3.475 & 2.111 \\
    8 & 0.322 &  2.439 &  2.020 &  3.411 & 1.944 \\
    9 & 0.322 &  2.228 &  2.277 &  3.382 & 1.721 \\
    10 & 0.279 &  1.640 &  2.468 &  3.219 & 1.215 \\
    }\MainVocabFreqDistribution

\pgfplotstableread[row sep=\\,col sep=&]{
    Position & XSum & MNLI & SST & STS & RTE & QNLI & QQP & MRPC\\
    1 & 4992 & 4936 & 3096 & 4319 & 4646 & 4946 & 4981 & 4449 \\
    2 & 4992 & 4862 & 2173 & 2851 & 3614 & 4834 & 4948 & 3061 \\
    3 & 4980 & 4740 & 1753 & 1911 & 2524 & 4671 & 4850 & 2041 \\
    4 & 4981 & 4623 & 1457 & 1413 & 1980 & 4562 & 4738 & 1454 \\
    5 & 4968 & 4560 & 1311 & 1163 & 1512 & 4402 & 4619 & 1149 \\
    6 & 4959 & 4473 & 1283 & 881  & 1207 & 4215 & 4567 & 902 \\
    7 & 4942 & 4304 & 1239 & 811  & 996  & 4013 & 4468 & 660 \\
    8 & 4912 & 4044 & 989  & 597  & 783  & 3755 & 4324 & 561 \\
    9 & 4825 & 3718 & 776  & 450  & 565  & 3402 & 4149 & 367 \\
    10& 3318 & 2015 & 330  & 162  & 160  & 1919 & 2913 & 119 \\
    }\AppenVocabDistribution

\pgfplotstableread[row sep=\\,col sep=&]{
    Position & MNLI & SST	& STS	& RTE &	QNLI &	QQP &	XSum & MRPC \\      
    1 & 3.376 & 2.256 & 1.411 & 1.457 & 2.885 & 3.200 & 4.231 & 1.535 \\
    2 & 2.294 & 1.486 & 0.645 & 0.560 & 1.900 & 2.242 & 3.141 & 0.613 \\
    3 & 2.011 & 1.359 & 0.522 & 0.411 & 1.670 & 1.854 & 2.831 & 0.477 \\
    4 & 1.822 & 1.249 & 0.458 & 0.356 & 1.482 & 1.591 & 2.623 & 0.415 \\
    5 & 1.702 & 1.163 & 0.390 & 0.322 & 1.354 & 1.427 & 2.466 & 0.380\\
    6 & 1.606 & 1.141 & 0.395 & 0.294 & 1.275 & 1.427 & 2.329 & 0.380 \\
    7 & 1.533 & 1.093 & 0.359 & 0.257 & 1.254 & 1.433 & 2.111 & 0.342 \\
    8 & 1.585 & 1.074 & 0.414 & 0.275 & 1.186 & 1.260 & 1.944 & 0.322 \\
    9 & 1.413 & 0.965 & 0.435 & 0.236 & 1.154 & 1.248 & 1.721 & 0.322 \\
    10& 1.137 & 0.927 & 0.268 & 0.209 & 1.056 & 1.192 & 1.215 & 0.279 \\
    }\AppendixVocabFreqDistribution

\pgfplotstableread[row sep=\\,col sep=&]{
    Position  & count &  ratio \\
      1 & 10490 & 0.208693922 \\
      2 & 3276 & 0.273868497 \\
      3 & 2879 & 0.331144932 \\
      4 & 3138 & 0.393574057 \\
      5 & 4056 & 0.474266388 \\
      6 & 4993 & 0.57359992 \\
      7 & 4944 & 0.671958619 \\
      8 & 4500 & 0.761484134 \\
      9 & 3627 & 0.833641699 \\
      10 & 2948 & 0.892290858 \\
      11 & 2301 & 0.938068238 \\
      12 & 1484 & 0.967591764 \\
      13 & 909  & 0.985675918 \\
      14 & 392  & 0.993474585 \\
      15 & 134  & 0.996140456 \\
      16 & 77 & 0.997672337 \\
      17 & 38 & 0.99842833 \\
      18 & 35 & 0.999124639 \\
      19 & 21 & 0.999542425 \\
    }\MSERatioDistributionRoBERTA
    
\pgfplotstableread[row sep=\\,col sep=&]{
Label &  series1 & series2 & series3 & series4 & series5 & series6 \\
A &  3410 &  2207 &  1418 &  1220 &  1997 &  238 \\
B &  1894 &  1322 &  1244 &  1035 &  660 &   0 \\
C &  2066 &  1418 &  1249 &  1354 &  1101 &  6 \\
D &  2022 &  3031 &  3121 &  3198 &  3050 &  15 \\
E &  608 &   2022 &  2968 &  3193 &  3192 &  6 \\
    }\MSEStackDistributionRoBERTA

\pgfplotstableread[row sep=\\,col sep=&]{
    Position  & count &  ratio \\
    1   & 17216  & 0.342504725 \\
    2   & 4789    & 0.437779767 \\
    3   & 4505    & 0.527404755 \\
    4   & 5538    & 0.637580822 \\
    5   & 4884    & 0.734745847 \\
    6   & 3675    & 0.807858351 \\
    7   & 3338    & 0.874266388 \\
    8   & 3020    & 0.934347956 \\
    9   & 1720    & 0.968566597 \\
    10  & 706 & 0.982612156 \\
    11  & 353 & 0.989634935 \\
    12  & 234 & 0.994290262 \\
    13  & 153 & 0.997334129 \\
    14  & 63  & 0.998587486 \\
    15  & 44  & 0.999462847 \\
    16  & 17  & 0.999801054 \\
    17  & 7   & 0.999940316 \\
    18  & 2   & 0.999980105 \\
    19  & 1   & 1 \\
    }\MSERatioDistribution
    
\pgfplotstableread[row sep=\\,col sep=&]{
Label &  series1 & series2 & series3 & series4 & series5 & series6 \\
A  & 4394  &  3598  &  3030  &  2817  &  3139  &  238 \\
B  & 2216  &  1696  &  1749  &  1873  &  1752  &  8 \\
C  & 1441  &  2055  &  2276  &  2534  &  2105  &  11 \\
D  & 1741  &  2127  &  2110  &  2000  &  2050  &  5 \\
E  & 208 & 524 & 835 & 776 & 954 & 3 \\
    }\MSEStackDistribution

\section{Isotropy Metric and DelDirection}
\label{appendix:isotropy_metric}

\textbf{Isotropy Metric.} \textcolor{black}{$\mathbf{E}$ denotes the pre-trained token embedding matrix of the PLM.}
Following previous studies~\citep{mu2018allbutthetop,bis-etal-2021-much,yu-etal-2022-rare}, we compute the isotropy of $\mathbf{E}$ using Eq.(1) from \citet{mu2018allbutthetop}, which is given by:
\begin{equation}
    I(\mathbf{E}) = \frac{\min_{\mathbf{b} \in \mathcal{B}} Z(\mathbf{b}) }{\max_{\mathbf{b} \in \mathcal{B}} Z(\mathbf{b})}, \nonumber
\end{equation}
\textcolor{black}{where $Z(\mathbf{b})$ is approximately constant, $\mathcal{B}$ is the set of eigenvectors of $\mathbf{E}^{T}\mathbf{E}$ with $T$ represents transposition operation.
}\newline
\newline
\noindent \textbf{DelDirection.} We regard DelDirection as our baseline since it is also a post-processing technique and can be applied to token-level vocabularies to improve isotropy for $\mathbf{E}$. DelDirection eliminates the common mean vector and top-$\beta$ dominating directions from $\mathbf{E}$.
According to~\citet{mu2018allbutthetop}, $\beta \approx h_{e}$/100, therefore we set $\beta=10$. 

\section{Downstream Datasets}
\label{appendix:vocab_distribution}

\begin{figure}[h]
\centering
\begin{subfigure}[t]{1\linewidth}
\begin{adjustbox}{width=1\columnwidth}
\begin{tikzpicture}
    \pgfplotsset{set layers}
    \begin{axis}[
            scale only axis,
            legend pos=outer north east,
            x post scale=2,
            y post scale=0.5,
            ylabel=\# of tokens,
            xlabel=Vocab bin,
            ymin=0, ymax=5500,
            xmin=1, xmax=10,
            symbolic x coords={1,2,3,4,5,6,7,8,9,10},
            xtick=data,
            ybar=2*\pgflinewidth,
            bar width=5pt,
            enlarge x limits=0.1,
            font=\huge,
            grid=major,
        ]
        \addlegendimage{/pgfplots/refstyle=plot_blue}\addlegendentry{Billsum}
        \addlegendimage{/pgfplots/refstyle=plot_green}\addlegendentry{XSum}
        \addlegendimage{/pgfplots/refstyle=plot_olive}\addlegendentry{CNNDM}
        \addlegendimage{/pgfplots/refstyle=plot_red}\addlegendentry{Y-BIGPATENT}
        \addplot+[fill=blue!40!white,draw=blue] table[x=Position,y=Billsum]{\MainVocabDistribution};\label{plot_blue}
        \addplot+[fill=green!40!white,draw=green] table[x=Position,y=XSum]{\MainVocabDistribution};\label{plot_green}
        \addplot+[fill=olive!40!white,draw=olive] table[x=Position,y=CNNDM]{\MainVocabDistribution};\label{plot_olive}
        \addplot+[fill=red!40!white,draw=red] table[x=Position,y=BIGPATENT]{\MainVocabDistribution};\label{plot_red}
    \end{axis}
\end{tikzpicture}
\end{adjustbox}
\caption{Number of tokens appearing in vocabulary $\mathcal{V}$.}
\label{fig:number_of_appeared_vocabs}
\end{subfigure}

\begin{subfigure}[t]{0.8\linewidth}
\begin{adjustbox}{width=1\columnwidth}
\begin{tikzpicture}
    \pgfplotsset{set layers}
    \begin{axis}[
            scale only axis,
            legend pos=outer north east,
            x post scale=1,
            y post scale=0.5,
            ylabel=Log(frequency),
            xlabel=Vocab bin,
            ymin=1, ymax=5,
            xmin=1, xmax=10,
            symbolic x coords={1,2,3,4,5,6,7,8,9,10},
            font=\huge,
        ]
        \addlegendimage{/pgfplots/refstyle=plot_billsum_freq}\addlegendentry{Billsum}
        \addlegendimage{/pgfplots/refstyle=plot_xsum_freq}\addlegendentry{XSum}
        \addlegendimage{/pgfplots/refstyle=plot_cnndm_freq}\addlegendentry{CNNDM}
        \addlegendimage{/pgfplots/refstyle=plot_bigpatent_freq}\addlegendentry{Y-BIGPATENT}
        \addplot+[sharp plot, mark=otimes, blue] table[x=Position,y=Billsum]{\MainVocabFreqDistribution};\label{plot_billsum_freq}
        \addplot+[sharp plot, mark=triangle, green] table[x=Position,y=XSum]{\MainVocabFreqDistribution};\label{plot_xsum_freq}
        \addplot+[sharp plot, mark=square, olive] table[x=Position,y=CNNDM]{\MainVocabFreqDistribution};\label{plot_cnndm_freq}
        \addplot+[sharp plot, mark=diamond, red] table[x=Position,y=BIGPATENT]{\MainVocabFreqDistribution};\label{plot_bigpatent_freq}
    \end{axis}
\end{tikzpicture}
\end{adjustbox}
\caption{Logarithmic averaged frequency of tokens appearing in the corresponding training set.}
\label{fig:averaged_logarithmic_token_frequency}
\end{subfigure}
\caption{Distribution of the BART vocabulary $\mathcal{V}$ in the training sets of text summarization datasets, considering both source and target tokens. The first 50,000 tokens in $\mathcal{V}$ are grouped into bins of 5,000 according to their index in $\mathcal{V}$ (\textit{e.g.,} [0:4,999]).
} 
\label{fig:main_vocab_distributions}
\end{figure}
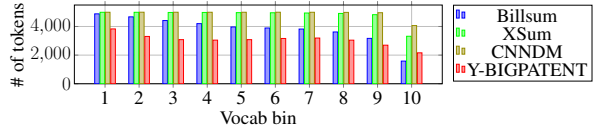
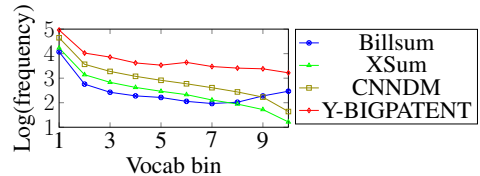

\begin{figure}[h]
\begin{subfigure}[t]{0.8\linewidth} 
\begin{adjustbox}{width=1\columnwidth}
\begin{tikzpicture}
    \pgfplotsset{set layers}
    \begin{axis}[
            scale only axis,
            x post scale=2,
            y post scale=1,
            ylabel=\# of tokens,
            xlabel=Vocab bin,
            ymin=0, ymax=5500,
            xmin=1, xmax=10,
            symbolic x coords={1,2,3,4,5,6,7,8,9,10},
            xtick=data,
            font=\huge,
        ]
        \addplot+[sharp plot, mark=x, mark options={solid}, dotted, blue] table[x=Position,y=MRPC]{\AppenVocabDistribution};\label{plot_mrpc}
        \addplot+[sharp plot, mark=+, mark options={solid}, dotted, green] table[x=Position,y=MNLI]{\AppenVocabDistribution};\label{plot_mnli}
        \addplot+[sharp plot, mark=diamond, mark options={solid}, olive, dashed]  table[x=Position,y=SST]{\AppenVocabDistribution};\label{plot_sst}
        \addplot+[sharp plot, mark=o, mark options={solid}, red] table[x=Position,y=STS]{\AppenVocabDistribution};\label{plot_sts}
        \addplot+[sharp plot, mark=square, black] table[x=Position,y=RTE]{\AppenVocabDistribution};\label{plot_rte}
        \addplot+[sharp plot, mark=o, mark options={solid}, dashed, purple] table[x=Position,y=QNLI]{\AppenVocabDistribution};\label{plot_qnli}
        \addplot+[sharp plot, mark=square, cyan] table[x=Position,y=QQP]{\AppenVocabDistribution};\label{plot_qqp}

    \end{axis}
\end{tikzpicture}
\end{adjustbox}
\caption{Number of tokens appearing in $\mathcal{V}$.} 
\end{subfigure}

\begin{subfigure}[h]{1\linewidth}  
\begin{adjustbox}{width=1\columnwidth}
\begin{tikzpicture}
    \pgfplotsset{set layers}
    \begin{axis}[
            scale only axis,
            legend pos=outer north east,
            x post scale=1,
            y post scale=0.5,
            ylabel=Log(frequency),
            xlabel=Vocab bin,
            ymin=0, ymax=3.5,
            xmin=1, xmax=10,
            symbolic x coords={1,2,3,4,5,6,7,8,9,10},
            font=\large,
        ]
        \addlegendimage{/pgfplots/refstyle=plot_mrpc_freq}\addlegendentry{MRPC}
        \addlegendimage{/pgfplots/refstyle=plot_mnli_freq}\addlegendentry{MNLI}
        \addlegendimage{/pgfplots/refstyle=plot_qqp_freq}\addlegendentry{QQP}
        \addlegendimage{/pgfplots/refstyle=plot_qnli_freq}\addlegendentry{QNLI}
        \addlegendimage{/pgfplots/refstyle=plot_sst_freq}\addlegendentry{SST}
        \addlegendimage{/pgfplots/refstyle=plot_sts_freq}\addlegendentry{STS}
        \addlegendimage{/pgfplots/refstyle=plot_rte_freq}\addlegendentry{RTE}
        \addplot+[sharp plot, mark=x, mark options={solid}, dotted, blue] table[x=Position,y=MRPC]{\AppendixVocabFreqDistribution};\label{plot_mrpc_freq}
        \addplot+[sharp plot, mark=+, mark options={solid}, dotted, green] table[x=Position,y=MNLI]{\AppendixVocabFreqDistribution};\label{plot_mnli_freq}
        \addplot+[sharp plot, mark=diamond, mark options={solid}, olive, dashed]  table[x=Position,y=SST]{\AppendixVocabFreqDistribution};\label{plot_sst_freq}
        \addplot+[sharp plot, mark=o, mark options={solid}, red] table[x=Position,y=STS]{\AppendixVocabFreqDistribution};\label{plot_sts_freq}
        \addplot+[sharp plot, mark=square, black] table[x=Position,y=RTE]{\AppendixVocabFreqDistribution};\label{plot_rte_freq}
        \addplot+[sharp plot, mark=o, mark options={solid}, dashed, purple] table[x=Position,y=QNLI]{\AppendixVocabFreqDistribution};\label{plot_qnli_freq}
        \addplot+[sharp plot, mark=square, cyan] table[x=Position,y=QQP]{\AppendixVocabFreqDistribution};\label{plot_qqp_freq}
    \end{axis}
\end{tikzpicture}
\end{adjustbox}
\caption{Logarithmic averaged frequency of tokens appearing in the corresponding training set.}
\end{subfigure}
\caption{Distribution of the BART vocabulary in the training sets of GLUE datasets. 
}
\label{fig:appendix_vocab_distributions}
\end{figure}
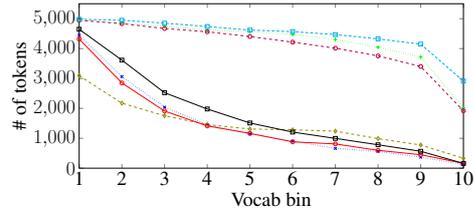
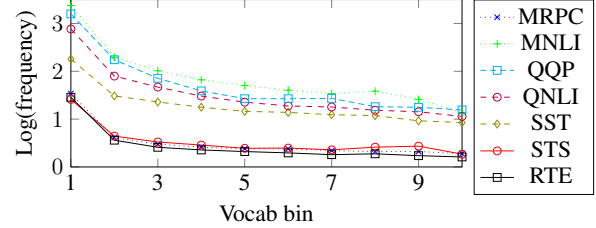

\noindent \textbf{Dataset Description.} Table~\ref{table:dataset_sizes} provides an overview of the data statistics for each task. 
For the GLUE task, we utilize Stanford Sentiment Treebank (SST), Microsoft Research Paraphrase Corpus (MRPC), Semantic Textual Similarity Benchmark (STS), Quora Question Pairs (QQP), MultiNLI (MNLI), Question NLI (QNLI), and Recognizing Textual Entailment (RTE) datasets.
For the text summarization task, CNNDM~\cite{NIPS2015_afdec700} comprises articles from CNN and Daily Mail newspapers, while Xsum~\citep{narayan-etal-2018-dont} consists of BBC articles paired with single-sentence summaries. BillSum~\citep{kornilova-eidelman-2019-billsum} contains summaries of US Congressional and California state bills. Y-BIGPATENT is the ``y'' category of BIGPATENT~\citep{sharma-etal-2019-bigpatent} and contains U.S. patent documents covering new or cross-sectional technology.\newline
\newline
\noindent \textbf{Dataset Distribution.} Figures~\ref{fig:main_vocab_distributions} and~\ref{fig:appendix_vocab_distributions} show the distribution of tokens appearing in the GLUE and text summarization datasets.
We observe that CNNDM dataset has the most uniform distribution and the greatest variety of appearing tokens among CNNDM, Y-BIGPATENT, XSum, and Billsum text summarization datasets.
The token frequency in Y-BIGPATENT dataset is significantly higher than in the other datasets, despite having the smallest variety of appearing tokens.
As these two figures show, the token frequency in text summarization datasets is much higher than that in the GLUE datasets. 
These findings suggest a potential difference when fine-tuning embeddings with respect to the task requirements. 

\begin{table*}[t]
\centering
\footnotesize
\resizebox{1\linewidth}{!}{
\begin{tabular}{llccc}
\toprule
\multirow{2}{*}{\textbf{Task}}                               & \multirow{2}{*}{\textbf{Dataset}}                                                                                                    & \multirow{2}{*}{\textbf{\# of train}} & \multirow{2}{*}{\textbf{\,\,\,\,\,\,\,\# of validation}} & \multirow{2}{*}{\textbf{\quad \# of test}} \\
& & & & \\
\midrule
\midrule
\multirow{4}{*}{Word similarity}  & RG65~\citep{10.1145/365628.365657} & - & - & 65 \\
 & SimLex-999~\citep{hill-etal-2015-simlex} & - & - & 999 \\
 & RW~\citep{luong-etal-2013-better} & - & - & 2,034 \\
 & SimVerb-3500~\citep{gerz-etal-2016-simverb} & - & - & 3,500 \\
\midrule
\multirow{7}{*}{GLUE}          & RTE                                                                       & 2, 490                          & 277                                  & 3, 000                         \\

                                    & MRPC                                                              & 3, 668                          & 408                                  & 1, 725                         \\
                                    & STS                                                            & 5, 749                          & 1, 500                               & 1, 379                         \\
               & SST                                                                 & 67, 349                         & 872                                  & 1, 821                         \\
                                    & QNLI                                                                                        & 104, 743                        & 5, 463                               & 5, 463                         \\

                                    & QQP                                                                                 & 363, 846                        & 40, 430                              & 390, 965                       \\
                                    & MNLI                                                                                            & 392, 702                        & 9, 815 (m) + 9, 832 (mm)                     & 9, 796 (m) + 9, 847 (mm)               \\
\midrule
\multirow{4}{*}{Text summarization} & BillSum                                  & 17, 054                         & 1, 895                               & 3, 269                         \\
                                    & Y-BIGPATENT  & 124, 397                        & 6, 911                               & 6, 911 \\
                                    & XSum                                                                              & 204, 045                        & 11, 332                              & 11, 334                        \\

                                    & CNNDM                                                                                  & 287, 227                        & 13, 368                              & 11,490                         \\
\bottomrule
\end{tabular}
}
\caption{Detailed statistic of train, validation and test datasets.
\textcolor{black}{For the MNLI dataset, we report Matched (m) and Mismatched (mm) sets.}
}
\label{table:dataset_sizes}
\end{table*}

\newpage

\section{Geometric Distribution of Embeddings}
\label{appendix:projected_embeddings}

\begin{figure}[h]
    \centering
    \includegraphics[width=1\linewidth]{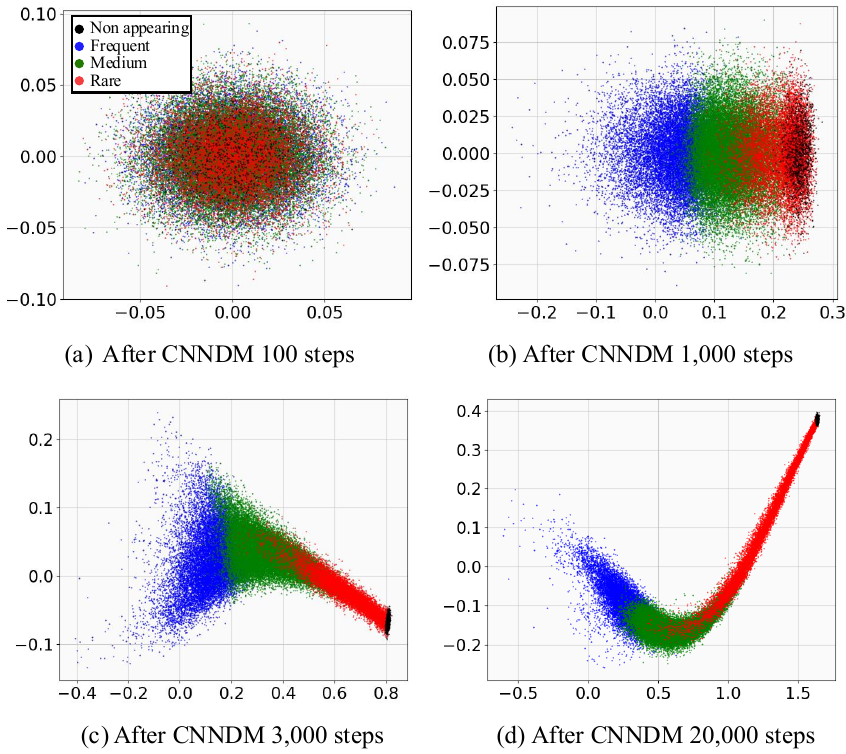}

\caption{Projected token embeddings of a randomly initialized model (BART architecture) training on the CNNDM dataset. 30\%, 50\%, and 20\% of the appearing tokens in BART's vocabulary $\mathcal{V}$ are assigned to the frequent, medium, and rare groups, respectively, based on their frequency in the training set. 
}
\label{fig:transformer_cnndm_procedure}
\end{figure}

\noindent \textbf{Fragile Embeddings without Pre-training.} To investigate whether pre-training helps models to be robust against degeneration into a cone shape, we compare pre-trained BART with a randomly initialized model (BART architecture). As shown in Figure~\ref{fig:transformer_cnndm_procedure}, after 1,000 training steps, the initialized model exhibits obvious degeneration, and after 3,000 steps, the shape of its embeddings becomes more like a narrow cone. This result aligns well with previous studies~\cite{gao2018representation,yu-etal-2022-rare}. In contrast, the pre-trained BART does not show such degeneration (Figure~\ref{fig:embedding_dynamics}). \newline
\newline
\noindent \textbf{Degeneration with More Updating.} Figures~\ref{figure:project_embedding_xsum_billsum} (e) and (g) show that before fine-tuning, the embeddings of DelDirection form a narrow ellipse. After fine-tuning, the minor axis of the ellipse lengthens, and the frequent, medium, and rare groups gradually disperse, eventually forming a square, as shown in Figures~\ref{figure:project_embedding_xsum_billsum} (f) and (h).
From Figures~\ref{figure:project_embedding_xsum_billsum} (f) and (h), we also observe that the more updating steps, the longer the original minor axis of the ellipse, and the higher $I(\mathbf{E})$ achieves by DelDirection. However, if DelDirection is fine-tuned further, the shape may resemble that in Figure~\ref{fig:embedding_dynamics} (h), resulting in a much smaller $I(\mathbf{E})$.
Comparing Figures~\ref{figure:project_embedding_xsum_billsum} (a) to (d) with Figures~\ref{figure:project_embedding_xsum_billsum} (i) to ($\ell$), we observe that BART + DefinitionEMB performs similarly to BART. Specifically, on the XSum dataset, there is minimal drift in embeddings from ``before'' to ``after'' fine-tuning. Conversely, for the Billsum dataset, after fine-tuning, the embeddings of two ellipses move closer together.\newline
\begin{figure*}[t]
    \centering
    \includegraphics[width=0.9\linewidth]{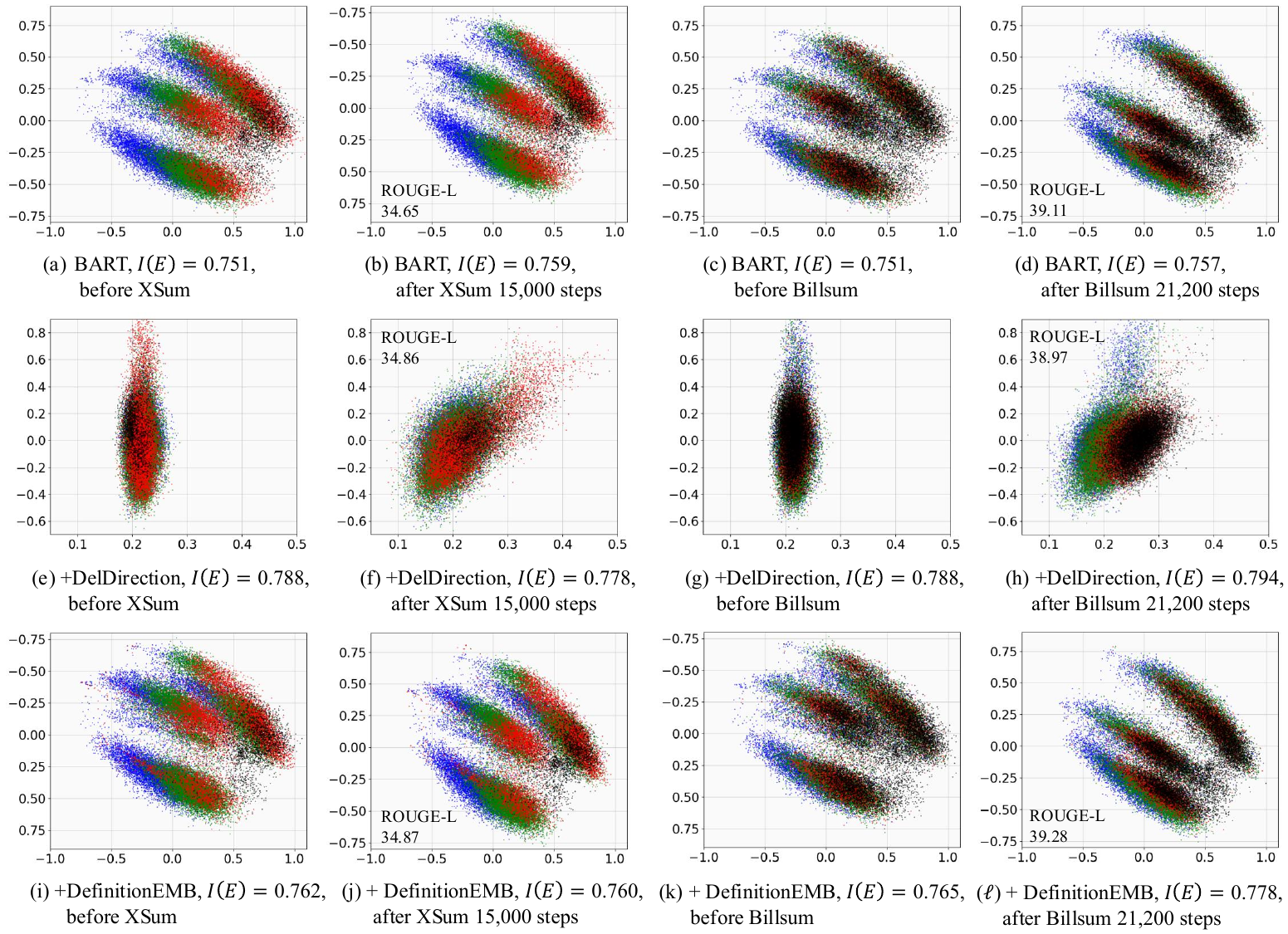}
\caption{Projected token embeddings of BART-related models before to after fine-tuning on the XSum and Billsum datasets. 30\%, 50\%, and 20\% of tokens appearing in $\mathcal{V}$ are assigned to the frequent, medium, and rare groups, respectively, based on their frequency in the fine-tuning set.
}
\label{figure:project_embedding_xsum_billsum}
\end{figure*}

\noindent \textbf{Architecture-Irrelevant Robustness.} We do not observe drift for BART-related models on the QQP and RTE datasets, as shown in Figure~\ref{figure:project_embedding_bart_glue}, which is different from text summarization datasets. This may be because the BART for classification tasks does not use the weight tying technique, and the token frequencies in QQP and RTE are much lower than in text summarization datasets.
Using DelDefinition for RoBERTa shows the spread of popular tokens from the original center, as shown in Figures~\ref{figure:project_embedding_roberta_glue} (e) and (f).
Although RoBERTa and BART have different model architectures (encoder-only vs. encoder-decoder), scales, and pre-training strategies, they both show similar robustness against representation degeneration. This suggests that PLMs' robustness against representation degeneration might not be directly related to these variables. \newline
\newline
\noindent \textbf{Case Studies.}
Figures \ref{figure:example_Wikimedia}, \ref{figure:example_clipping}, and~\ref{figure:example_409} show case studies of specific tokens before and after replacing their token embeddings. 
Non-appearing, rare, medium, and frequent groups in the CNNDM dataset are represented by black, red, green, and blue points respectively.
In BART, the central tokens are surrounded by tokens of the same frequency, rather than those with related semantics. In the case of BART+DelDirection, we observe tokens with different token frequencies surrounding the central word. However, using DelDirection also does not guarantee semantically related neighbors. After replacing embeddings with DefinitionEMB, semantically related tokens appear in the surrounding of the central tokens.

\begin{figure*}[t]
    \centering
    \includegraphics[width=0.9\linewidth]{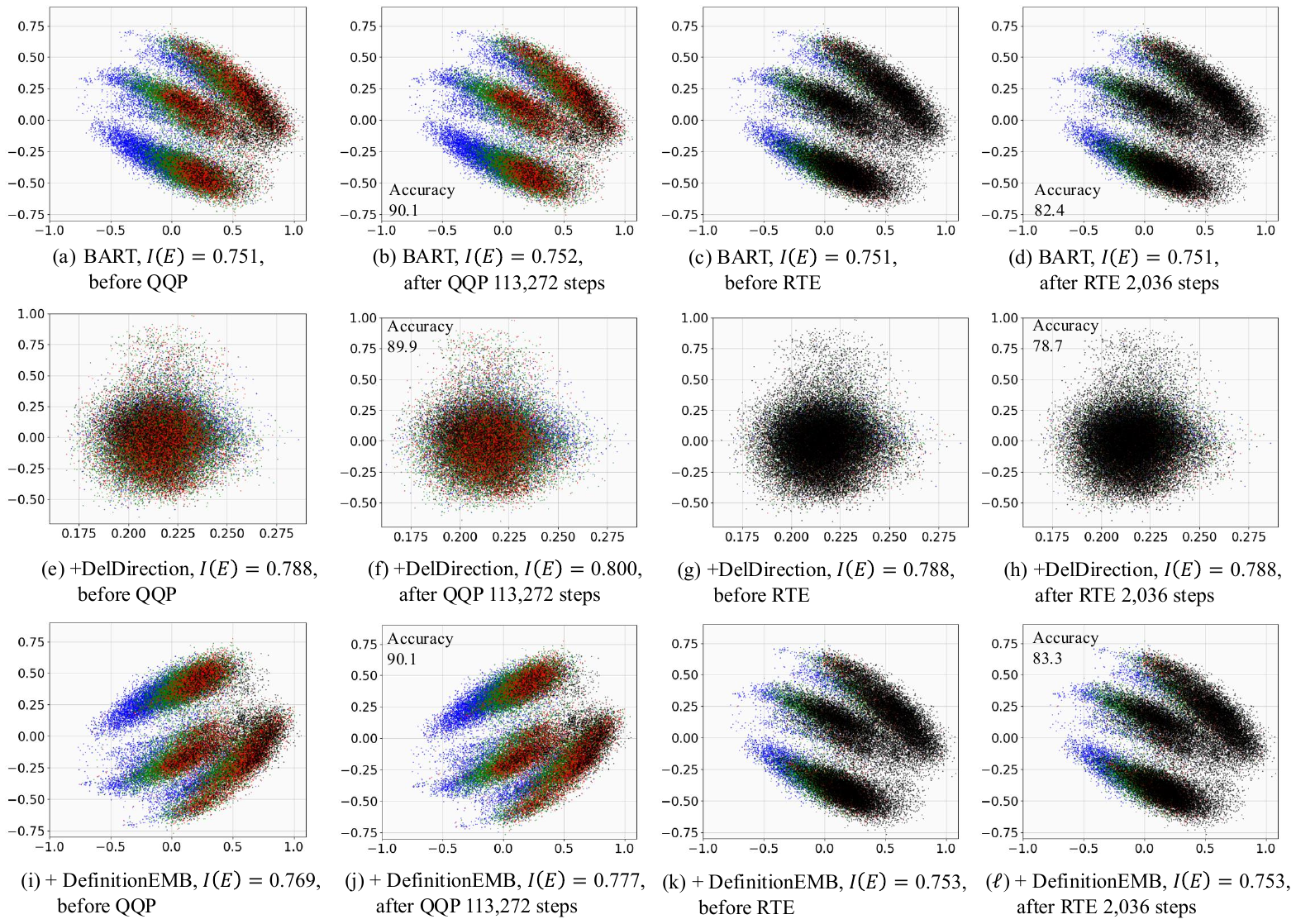}
\caption{Projected token embeddings of BART-related models ``before'' and ``after'' fine-tuning on the QQP and RTE datasets.}
\label{figure:project_embedding_bart_glue}
\end{figure*}

\begin{figure*}[t]
    \centering
    \includegraphics[width=0.9\linewidth]{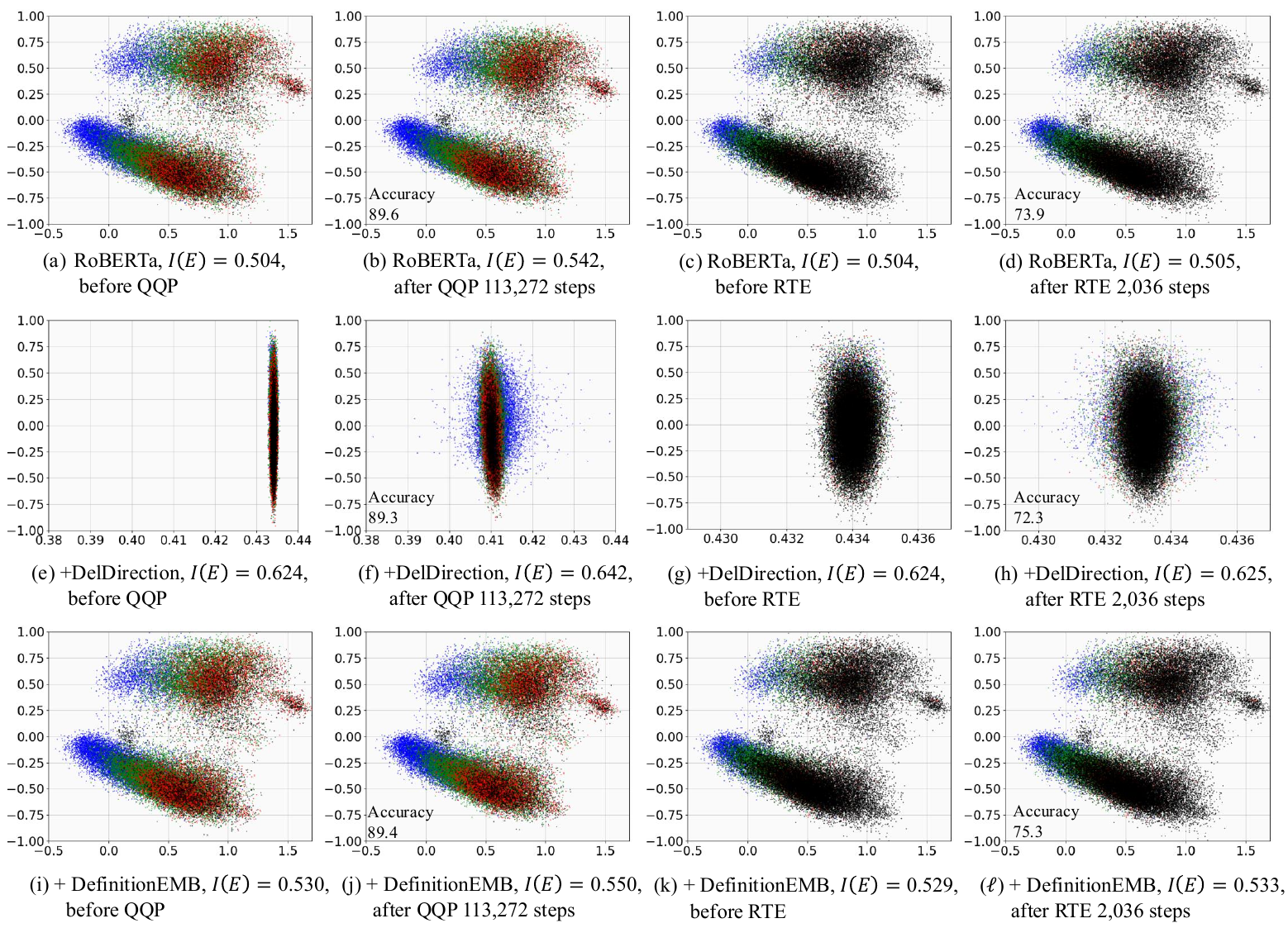}

\caption{Projected token embeddings of RoBERTa-related models ``before'' and ``after'' fine-tuning on the QQP and RTE datasets.}
\label{figure:project_embedding_roberta_glue}
\end{figure*}

\begin{figure*}[t]
    \centering
    \includegraphics[width=1\linewidth]{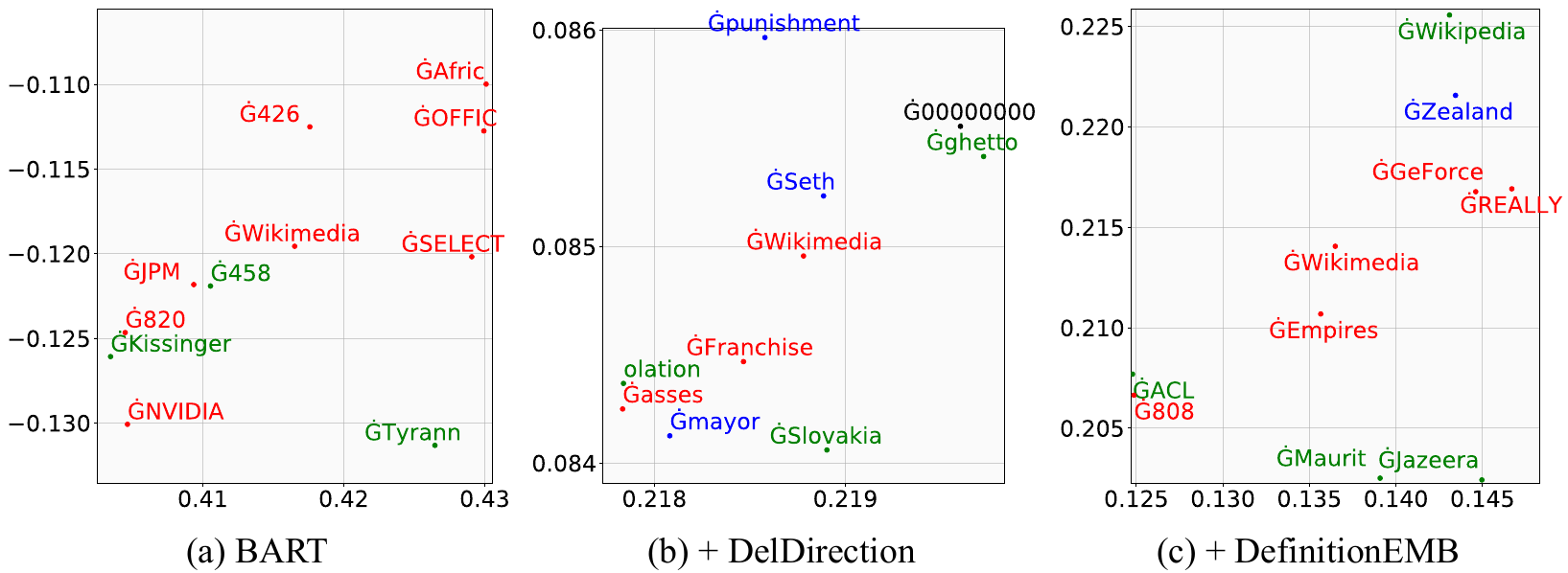}

\caption{\textcolor{black}{Case study of the token embeddings of the token ``ĠWikimedia'' and its surrounding tokens, where ``Ġ'' denotes whitespace. 
}}
\label{figure:example_Wikimedia}
\end{figure*}

\begin{figure*}[t]
    \centering
    \includegraphics[width=1\linewidth]{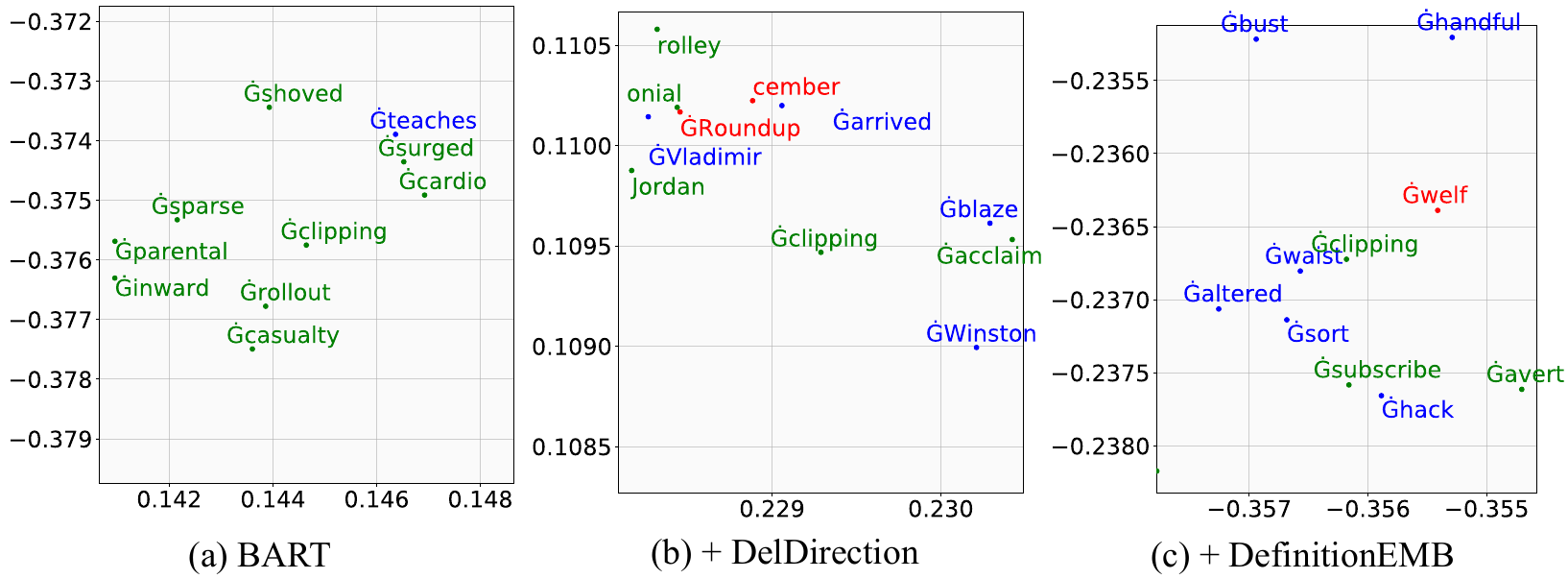}

\caption{\textcolor{black}{Case study of the token embeddings of the token ``Ġclipping'' and its surrounding tokens.}}
\label{figure:example_clipping}
\end{figure*}

\begin{figure*}[t]
    \centering
    \includegraphics[width=1\linewidth]{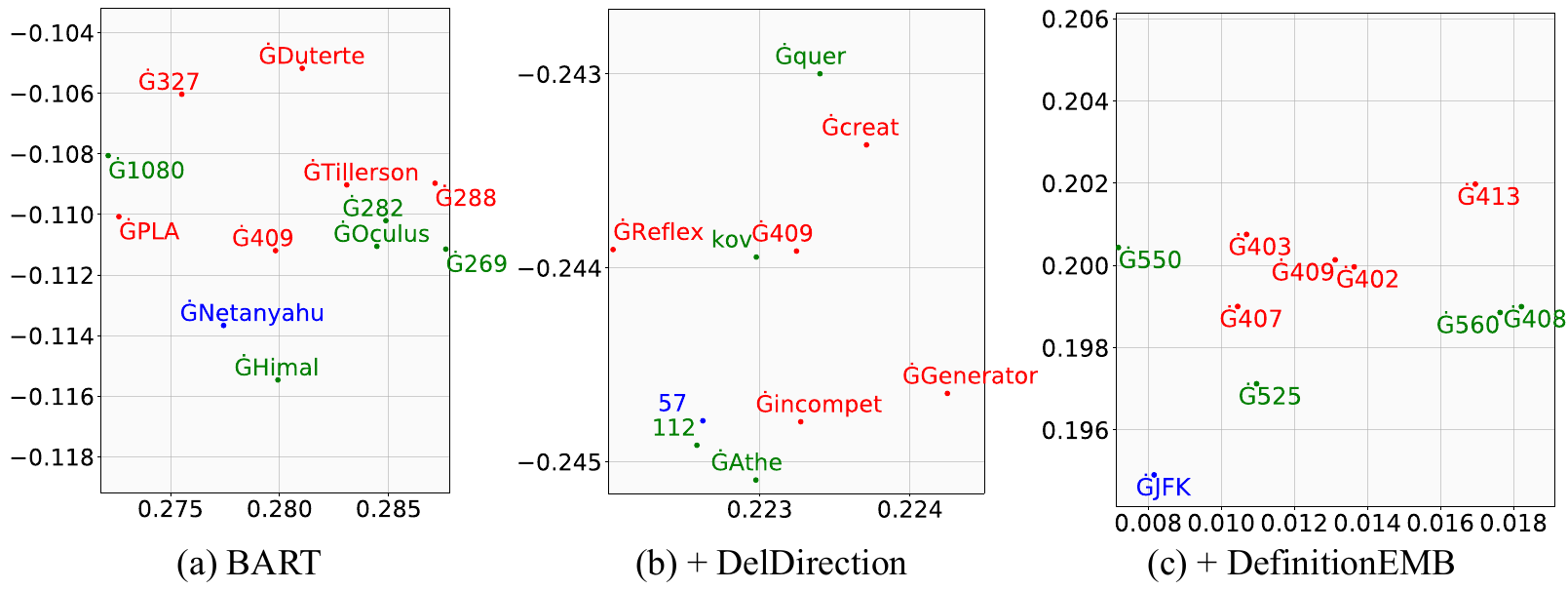}

\caption{\textcolor{black}{Case study of the token embeddings of the token ``Ġ409'' and its surrounding tokens.
}}
\label{figure:example_409}
\end{figure*}

\clearpage

\begin{table*}[t]
\resizebox{\textwidth}{!}{
\begin{tabular}{p{2.4cm}|p{15.5cm}}
\toprule
Query                       & Please help me to check the semantic relationship between the token ``everlasting'' and other tokens, including reassure, legalize, validate, underworld, typo, misdemeanor, jurisd, scripting, APIs, io, bye, complaining, LAB, arcade, incentive, hematically, mind, stagn, sun, vast, star, tw, general, click, rusty. Please return scores in the range [0,1] to indicate their semantic relationship, and tell me the reason. \\
\midrule
\multirow{5}{*}{ChatGPT 3.5} & validate: 0.3 - "Validate" can mean to confirm or verify the accuracy or truth of something, which could have a moderate semantic relationship with "everlasting" if used in a context related to eternal validity or truth.                                                                                                                                                                                                          \\
                             & mind: 0.2 - "Mind" refers to the faculty of consciousness and thought, which could have a moderate semantic relationship with "everlasting" if used in a context related to eternal consciousness or existence.                                                                                                                                                                                                                        \\
                             & sun: 0.3 - "Sun" refers to the star around which the Earth orbits and has a moderate semantic relationship with "everlasting" due to its association with eternal cycles and longevity.                                                                                                                                                                                                                                                \\
                             & vast: 0.3 - "Vast" refers to something extremely large or extensive, which could have a moderate semantic relationship with "everlasting" if used in a context related to eternal expansiveness.                                                                                                                                                                                                                                       \\
                             & star: 0.3 - "Star" refers to a celestial body that emits light and has a moderate semantic relationship with "everlasting" due to its association with eternal cycles and longevity.                                                                                                                                                                                                                                                   \\
\midrule
\multirow{5}{*}{Claude 3 Haiku}   & "everlasting" and "validate": Score: 0.4 Reason: "Everlasting" and "validate" share a sense of permanence and confirmation, as "validate" means to confirm or corroborate something as true or correct, which is somewhat related to the idea of something being eternal or lasting forever.                                                                                                                                           \\
                             & "everlasting" and "mind": Score: 0.4 Reason: "Everlasting" suggests something that lasts forever, while the "mind" is a key part of human consciousness and experience, which are somewhat related in the sense that the human mind may be seen as an eternal or enduring aspect of the self.                                                                                                                                          \\
                             & "everlasting" and "sun": Score: 0.4 Reason: "Everlasting" suggests something that lasts forever, and the "sun" is a celestial body that has existed for billions of years and is expected to continue shining for a very long time, so there is a connection between the idea of something being "everlasting" and the sun's enduring presence.                                                                                        \\
                             & "everlasting" and "vast": Score: 0.5 Reason: "Everlasting" suggests something that is immense or boundless in its duration, while "vast" describes something that is immense in size or scope, so there is a semantic overlap between the two terms in terms of conveying a sense of immensity or grandeur.                                                                                                                            \\
                             & "everlasting" and "star": Score: 0.4 Reason: Similar to the connection with the "sun", the "star" is another celestial body that exists for long durations, so the idea of something being "everlasting" is somewhat related to the enduring nature of stars.\\
\bottomrule                    
\end{tabular}
}
\caption{The query to ChatGPT 3.5 and Claude 3 Haiku, along with part of their responses.}
\label{table:llm_query}
\end{table*}

\begin{figure*}[t]
\centering
\begin{subfigure}[t]{1\linewidth}
\caption{\textbf{Full prompt}}
\centering
\begin{tikzpicture}
\tikzstyle{every node}=[font=\small,scale=0.9]
\node[align=left]{
The \_ definition \_ of \_ discomfort \_ is \_ To \_ cause \_ annoyance \_ or \_ distress \_ to \_ . \_ Its \_ part-of-speech \_ , \_ bpe-form \\
\_ without \_ space \_ , \_ capitalization \_ , \_ and \_ uppercase \_ are \_ verb \_ , discomfort \_ , \_ Discomfort \_ , \_ and \_ \\
DISCOMFORT \_ , \_ respectively \_ .
};
\end{tikzpicture}
\end{subfigure}

\begin{subfigure}[t]{1\linewidth}
\caption{\textbf{Source for the encoder-decoder PLM}}
\centering
\begin{tikzpicture}
\tikzstyle{every node}=[font=\small,scale=0.9]
\node[align=left]{
The \_ definition \_ of <$\text{MASK}_{1}$> \_ is \_ To \_ cause \_  annoyance \_ or \_ distress \_ to \_ . \_ Its \_ part-of-speech \_ , \_ bpe-form \\
\_ without \_ space \_ , \_ capitalization \_ , \_ and \_ uppercase \_ are \_ verb \_ , <$\text{MASK}_{2}$> \_ , <$\text{MASK}_{3}$> \_ , \_ and <$\text{MASK}_{4}$> \\
\_ , \_ respectively \_ .
};
\end{tikzpicture}
\end{subfigure}

\begin{subfigure}[t]{1\linewidth}
\caption{\textbf{Target for the encoder-decoder PLM}}
\centering
\begin{tikzpicture}
\tikzstyle{every node}=[font=\small,scale=0.9]
\node[align=left]{
<$\text{MASK}_{1}$> \_ discomfort <$\text{MASK}_{2}$> discomfort <$\text{MASK}_{3}$> \_ Discomfort <$\text{MASK}_{4}$> \_ DISCOMFORT
};
\end{tikzpicture}
\end{subfigure}

\begin{subfigure}[t]{1\linewidth}
\caption{\textbf{Source for the encoder-only PLM during training}}
\centering
\begin{tikzpicture}
\tikzstyle{every node}=[font=\small,scale=0.9]
\node[align=left]{
The \_ definition \_ of <$\text{MASK}$> \_ is \_ To \_ cause \_ annoyance \_ or \_ distress \_ to \_ . \_ Its \_ part-of-speech \_ , \_ \\
bpe-form \_ without \_ space \_ , \_ capitalization \_ , \_ and \_ uppercase \_ are \_ verb \_ , <$\text{MASK}$>comfort \_ , ĠDis <$\text{MASK}$> \\
\_ , \_ and <$\text{MASK}$> COM \textcolor{red}{\textit{BS}} \textcolor{red}{\textit{ĠNations}}\_ , \_ respectively \_ .
};
\end{tikzpicture}
\end{subfigure}

\begin{subfigure}[t]{1\linewidth}
\caption{\textbf{Source for the encoder-only PLM during inference}}
\centering
\begin{tikzpicture}
\tikzstyle{every node}=[font=\small,scale=0.9]
\node[align=left]{
The \_ definition \_ of <$\text{MASK}$> \_ is \_ To \_ cause \_ annoyance \_ or \_ distress \_ to \_ . \_ Its \_ part-of-speech \_ , \_ bpe-form \_ without \_\\
space \_ , \_ capitalization \_ , \_ and \_ uppercase \_ are \_ verb \_ , discomfort \_ , \_ Discomfort \_ , \_ and \_ DISCOMFORT \_ , \\
\_ respectively \_ .
};
\end{tikzpicture}
\end{subfigure}

\caption{Example of constructing prompts for the word ``discomfort''.
``\_'' is whitespace.
<$\text{MASK}_{i}$> denotes the $i^{\text{th}}$ mask token. \textcolor{red}{\textit{Italic}} indicates randomly replaced tokens.}
\label{fig:appendix_example_prompts}
\end{figure*}

\clearpage
\section{Semantically Related Tokens Recognized by Large Language Models}
\label{appendix:semantically_related_llm}

Table~\ref{table:llm_query} shows the query we used to ask ChatGPT 3.5 and Claude 3 Haiku about the semantic relationship between ``Ġeverlasting'' and other tokens, along with their corresponding responses for semantics-related tokens only.

\section{Settings for DefinitionEMB}
\subsection{Details of Cloze Exercise}
\label{appendix:corrupted_prompts}
\begin{figure}[h]
    \centering
    \includegraphics[width=\linewidth]{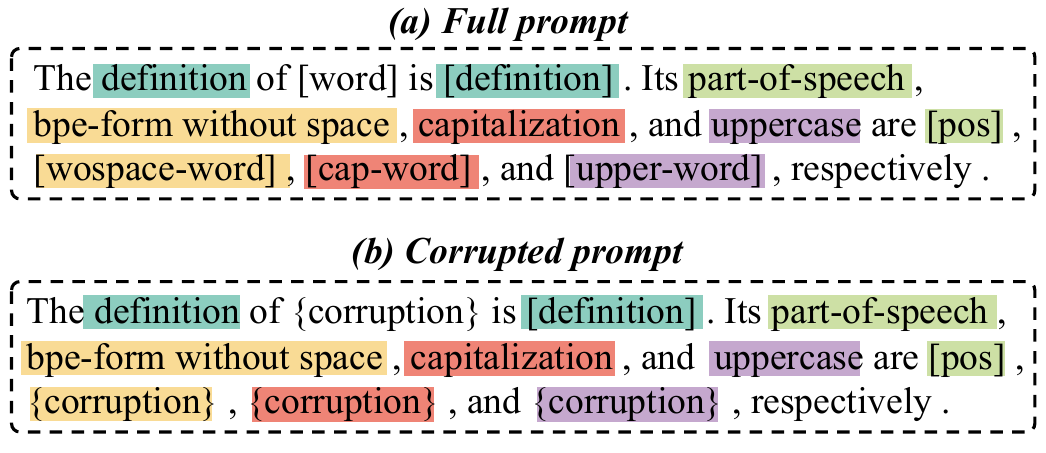}
\caption{Constructed prompts. 
Brackets [] are a placeholder for the given word and its corresponding information.
Texts with the same color indicate positions of a prompt and corresponding word information.
\{corruption\} indicates the span for corrupted tokens.
The bpe-form without space refers to the word's surface-form without the symbol ``Ġ'' when using the BART's tokenizer.
}
\label{fig:example_prompts}
\end{figure}

\noindent \textbf{Full Prompt.} As shown in Figure~\ref{fig:example_prompts} (a), to include complete definition for better re-construction, our full prompt incorporates the definition, part-of-speech, capitalization, and case sensitivity of the word $\boldsymbol{w}$, along with the tokenizer's specific settings. Figure~\ref{fig:appendix_example_prompts} (a) shows examples of full prompt for the word ``discomfort''. \newline
\newline
\noindent \textbf{Corruption Strategy.} As shown in Figures~\ref{fig:example_prompts} (b), parts of tokens within the corruption spans are randomly corrupted, as detailed below.

For encoder-only PLM, as shown in Figures~\ref{fig:appendix_example_prompts} (d-e), tokens within the corruption spans are corrupted using a BERT-style masking strategy. Specifically, during training, 50\% of the tokens are replaced with a special mask token <MASK>, 25\% are replaced with a random token, and the remaining 25\% are left unchanged.\footnote{Our pre-experiments demonstrate that a large mask ratio would result in slow convergence, while a small ratio would cause limited change between $\mathbf{e}(v)$ and $\tilde{\mathbf{e}}(v)$. To ensure computational efficiency and to prevent the model from relying too heavily on unmasked tokens, we manually set these ratios.} During inference, only one token within the corruption spans is replaced with <MASK> at a time, and this procedure is repeated until all tokens in the corruption spans have been replaced.

For the encoder-decoder PLM, as shown in Figures~\ref{fig:appendix_example_prompts} (b-c), we follow the T5-style masking approach by \citet{10.5555/3455716.3455856} to corrupt tokens. Specifically, we replace each corruption span with a mask token to construct the source sequence. Subsequently, we construct the target sequence using the replaced tokens delimited by the mask tokens used for replacement.\newline
\newline
\noindent \textbf{Embedding Reconstruction.} DefinitionEMB re-constructs embedding as follows:

For the encoder-only model, we re-construct embedding according to the masked language modeling~\cite{devlin-etal-2019-bert}. Let $v_k$ be a corrupted token, with $k \in \kappa$ indicating its position in the full prompt.
We utilize the corrupted prompt as input, where $v_k$ is replaced by <MASK>. And then, the last hidden states $\mathbf{s}_k$ is computed at position $k$ and mapped to the re-constructed embedding of $v_k$.

For the encoder-decoder model, we re-construct embedding according to the causal language modeling~\citep{hyndman2018forecasting}. The constructed target sequence is denoted as $\boldsymbol{y}=(y_1, \ldots, y_{2J})$, where $J$ is the number of corrupted tokens. For any $j \in [1, J]$, $y_{2j}$ and $y_{2j-1}$ denote corrupted tokens and their corresponding delimiters, respectively. Input the source and prefix target tokens $\boldsymbol{y}_{<2j}=(y_1, \ldots, y_{2j-1})$, we map the last hidden states of $y_{2j-1}$ to the re-constructed embedding of $y_{2j}$ in the decoder.

\clearpage

\subsection{Pre-experiment for Replacing Strategy}
\label{appendix:pre_experiment}
\begin{table}[h]
\centering
\begin{tabular}{lcc}
\toprule
\multirow{2}{*}{\textbf{Strategy}} & \multirow{2}{*}{\textbf{RoBERTa}} & \multirow{2}{*}{\textbf{BART}} \\
& &\\
\midrule
\midrule
None         &  87.5 &     87.8                              \\
\midrule
Random         &        87.4            &             87.3                      \\
Top    &    86.0   &  87.6                          \\
Last &   \textbf{87.7}  & \textbf{88.3}    \\
\bottomrule
\end{tabular}
\caption{Using three different strategies to replace tokens in $\mathcal{V}_{MRPC}$ with DefinitionEMB ($\alpha=5$) for RoBERTa-base and BART-large on the MRPC test set.}
\label{table:pre-experiment}
\end{table}

\noindent Considering the appearance bias as indicated in Figures~\ref{fig:embedding_dynamics} (e) and (j), we conduct pre-experiments to investigate which type of tokens in $\mathcal{V}_{[task]}$ should be replaced. DefinitionEMB replaced the embeddings of $\min(\alpha \% * |\mathcal{V}|, |\mathcal{V}_{[task]}|)$ of tokens in $\mathcal{V}_{[task]}$ using one of the following strategies:

\noindent \textbf{Random}: Randomly replace tokens. 

\noindent \textbf{Top}: Replace top tokens in $\mathcal{V}_{[task]}$, where index $\ge 5000$.

\noindent \textbf{Last}: Replace last tokens in $\mathcal{V}_{[task]}$.

As shown in Table~\ref{table:pre-experiment}, replacing the last tokens in $\mathcal{V}_{[task]}$ results in the highest accuracy.

\section{Hyperparameters}
\label{appendix:hyperparameters}

\begin{table}[ht]
\centering
\resizebox{0.8\linewidth}{!}{
\begin{tabular}{lrr}
\toprule
\multirow{2}{*}{\textbf{Hyperparameters}} & \multirow{2}{*}{\textbf{RoBERTa}} & \multirow{2}{*}{\textbf{BART}} \\
& & \\\midrule
\midrule
\# of updates & 400,000  &  250,000 \\
\# of warm-up updates &24,000 & 20,000 \\
\bottomrule
\end{tabular}
}
\caption{Hyperparameters used for training DefinitionEMB initialized from BART and RoBERTa.}
\label{table:train_hyperparameters}
\end{table}

\begin{table}[ht]
\centering
\resizebox{1\linewidth}{!}{
\begin{tabular}{lrrrrrrr}
\toprule
\multirow{2}{*}{\textbf{Hyperparameters}} & \multirow{2}{*}{\textbf{SST}} & \multirow{2}{*}{\textbf{MRPC}} & \multirow{2}{*}{\textbf{STS}} & \multirow{2}{*}{\textbf{QQP}} & \multirow{2}{*}{\textbf{MNLI}} & \multirow{2}{*}{\textbf{QNLI}} & \multirow{2}{*}{\textbf{RTE}} \\
& & & & & & & \\ \midrule
\midrule
\# of updates & 20,935  & 2,296  & 3,598 & 113,272 & 123,873  &  33,112 &  2,036 \\
\# of warm-up updates & 1,256 & 137 & 214 & 28,318  & 7,432 & 1,986 &  122 \\
Batch size (sentences) & 32 & 16 & 16 & 32 & 32  & 32 &  16 \\
Learning rate & 1e-5 & 1e-5 & 2e-5 & 1e-5 & 1e-05 & 1e-05 & 2e-05 \\
\bottomrule
\end{tabular}
}
\caption{Hyperparameters used for fine-tuning RoBERTa-related models across different datasets.}
\label{table:finetune_hyperparameters_roberta}
\end{table}

\begin{table*}[ht]
\centering
\resizebox{1\linewidth}{!}{
\begin{tabular}{ll}
\toprule
\multirow{2}{*}{\textbf{Used artifacts}} & \multirow{2}{*}{\textbf{Note}}                                                                      \\ 
& \\
\midrule
\midrule
RoBERTa        & \url{https://huggingface.co/roberta-base}                \\
BART           & \url{https://huggingface.co/facebook/bart-large}  \\
BART for GLUE & \url{https://github.com/facebookresearch/fairseq/blob/main/examples/bart/README.glue.md} \\
BART for CNNDM & \url{https://github.com/facebookresearch/fairseq/blob/main/examples/bart/README.summarization.md} \\
Wiktionary & \url{https://en.wiktionary.org/wiki/Wiktionary:Main_Page} \\
Wiktionary(extracted)         & \url{https://github.com/tatuylonen/wiktextract/tree/master}\\
Isotropy metric       & \url{https://github.com/danielbis/tooMuchInCommon/blob/main/src/isotropy.py}        \\
View contextual embedding & \url{https://github.com/TideDancer/iclr21_isotropy_contxt} \\
Files2rouge       & \url{https://github.com/pltrdy/files2rouge}        \\
Paired bootstrap resampling & \url{https://github.com/neubig/util-scripts/blob/master/paired-bootstrap.py} \\
Fairseq        & \url{https://github.com/facebookresearch/fairseq/} \\
HuggingFace & \url{https://github.com/huggingface/transformers/} \\
ChatGPT 3.5 & \url{https://chat.openai.com/} \\
Claude 3 Haiku & \url{https://claude.ai/chat/} \\
\bottomrule
\end{tabular}
}
\caption{Used artifacts.}
\label{table:used_artifacts}
\end{table*}
\begin{table*}[ht]
\centering
\resizebox{1\linewidth}{!}{
\begin{tabular}{lrrrrrrrrrrr}
\toprule
\multirow{2}{*}{\textbf{Hyperparameters}} & \multirow{2}{*}{\textbf{SST}} & \multirow{2}{*}{\textbf{MRPC}} & \multirow{2}{*}{\textbf{STS}} & \multirow{2}{*}{\textbf{QQP}} & \multirow{2}{*}{\textbf{MNLI}} & \multirow{2}{*}{\textbf{QNLI}} & \multirow{2}{*}{\textbf{RTE}} & \multirow{2}{*}{\textbf{CNNDM}}  & \multirow{2}{*}{\textbf{Y-BIGPATENT}} & \multirow{2}{*}{\textbf{XSUM}}  & \multirow{2}{*}{\textbf{Billsum}}\\ 
& & & & & & & & & & & \\ \midrule
\midrule
\# of updates & 7,150  & 700  & 1,800 & 113,920 &  43,210 &  33,290 & 1020 &  20,000 & 92,880  &  15,000 & 21,320 \\
\# of warm-up updates & 429 & 42 & 108 &  6,835 & 2,593 & 1,997 & 61 & 500 &  7,430 &  500 &  1,705 \\
Batch size (sentences) & 128 & 64 & 32 & 32 & 256  & 32 & 32 & - & -  &  - & - \\
Batch size (tokens) & -  & -  & -  & - &  - &  - & -  & 65,536 & 8,192  & 32,768  & 8,192 \\
Learning rate & 5e-6 & 2e-5 & 2e-5 & 1e-5 & 5e-6 & 1e-5 & 1e-5 & 3e-05 & 3e-5  & 3e-05  & 3e-5  \\
\bottomrule
\end{tabular}
}
\caption{Hyperparameters used for fine-tuning BART-related models across different datasets.}
\label{table:finetune_hyperparameters}
\end{table*}

\noindent \textbf{Used Artifacts.} Table~\ref{table:used_artifacts} lists the artifacts utilized in the study.  We use the Fairseq~\citep{ott-etal-2019-fairseq} and HuggingFace~\citep{wolf-etal-2020-transformers} to reproduce all models and run the downstream tasks.

We adhere to the original fine-tuning settings of RoBERTa and BART on the GLUE task and CNNDM dataset. BART utilizes the weight tying~\citep{press-wolf-2017-using} technique when predicting texts, which involves using $\mathbf{E}$ as the weight matrix for computing logits.
Details of the hyperparameter settings for training DefinitionEMB and fine-tuning models are outlined in Tables~\ref{table:train_hyperparameters}, ~\ref{table:finetune_hyperparameters_roberta}, and~\ref{table:finetune_hyperparameters}. During the training of DefinitionEMB, we utilize the Adam optimizer with a batch size of 4,096 tokens and a 0.0001 learning rate with an ``inverse square root'' schedule. For fine-tuning models on GLUE and text summarization tasks, we employ the Adam optimizer and utilized a ``polynomial decay'' learning rate schedule.
To reduce computation, we freeze the embedding layer of DefinitionEMB during training for BART. 
\textcolor{black}{As for the initialized model in Figure~\ref{fig:transformer_cnndm_procedure}, we set learning rate, batch size (tokens), number of updates, and number of warm-up steps as 0.001,  64,000, 50,000, and 4,000, respectively.} \newline
\newline
\noindent \textbf{Definitions for Numbers and Named Entities.} We add definitions for 1,252 numbers in PLM's vocabulary $\mathcal{V}$ by translating numbers into their corresponding words, such as ``2'' to ``two''. Furthermore, we added definitions for 136 named entity tokens in $\mathcal{V}$, such as ``ĠNVIDIA'', based on their Wikipedia pages or Google search results. These definitions are available at our anonymous Github \url{https://anonymous.4open.science/r/DefinitionEMB-2754}.
\newline
\newline
\noindent \textbf{Rare Token Subset.} To construct the rare token subset that used in Table~\ref{table:filtered_cnndm}, we first filter the CNNDM test set to include target sentences whose tokens all appear in the training set. Additionally, each filtered target sentence must contain at least 5\% rare tokens with indices larger than 40,000 in $\mathcal{V}$, and these tokens' embeddings can be replaced by DefinitionEMB. This process finally yield 65 pairs of data. This subset is also available at our anonymous Github.
\newline
\newline
\noindent \textbf{Tuning $\alpha$.} For each downstream task, we tune $\alpha$ with a single trial on the corresponding validation set.
Tables~\ref{table:tuning_mrpc} to~\ref{table:tuning_bigpatent} display the performance of Basline+DefinitonEMB with various $\alpha$.
Tables~\ref{table:tuned_alpha_glue} and~\ref{table:tuned_alpha_text_summarization} provide the tuned $\alpha$ for each downstream dataset.
Overall, datasets in the GLUE task exhibit smaller $\alpha$ than those in the text summarization task. This may be because the text summarization task involves a larger number of input tokens than the GLUE task.
Additionally, the text summarization task involves predicting tokens, whereas BART employs a weight tying technique to connect token embeddings for prediction, enabling token embeddings to be updated more during fine-tuning than in the GLUE task.
Among all datasets, the SST dataset has the smallest tuned $\alpha$, set at 1. This could be attributed to the dataset having the fewest unique tokens.
The STS, RTE, and MPRC datasets have a similar number of unique tokens, slightly higher than that of the SST dataset, resulting in $\alpha$ values of 3 and 5.
In the GLUE task, the QQP, QNLI, and MNLI datasets have the highest number of unique tokens, leading to $\alpha$ values for BART of 3, 5, and 10, respectively. 
However, tuned $\alpha$ values of the three datasets for RoBERTa are all set to 3: this may be because MSE is larger in RoBERTa than in BART as discussed in Appendix~\ref{appendix:mse_between_embeddings}.
Among the text summarization datasets, Billsum has the lowest token frequencies, resulting in the smallest $\alpha$ value of 7.
Despite having the highest token frequencies, the Y-BIGPATENT dataset has the lowest number of unique tokens, resulting in an $\alpha$ value of only 30.
The CNNDM dataset yields the highest $\alpha$ value, set at 100, possibly due to its most uniformly distributed and largest number of unique tokens, and larger training corpus.
Although the XSum dataset also contains a large number of unique tokens, it has a smaller training corpus than CNNDN, resulting in a smaller $\alpha$ value of 10.
These findings suggest a potential relationship between the distribution of training data and the value of $\alpha$. Specifically, datasets with a larger number of unique tokens, along with more training examples, tend to result in higher $\alpha$ values.
\clearpage
\begin{table}[t]
\centering
\scriptsize
\resizebox{1\linewidth}{!}{
\begin{tabular}{l|lllll}
\toprule
\diagbox{Baseline}{$\alpha$} &  1    & 5             & 10   & 20   \\
\midrule
\midrule
RoBERTa  &  89.2 & \textbf{90.2} & 89.5 & 89.7 \\
\midrule
BART     &  88.2 & \textbf{90.2}          & 88.7 & 89.2 \\
\bottomrule
\end{tabular}
}
\caption{Accuracy for Baseline+DefinitionEMB with various $\alpha$ on the MRPC validation set.}
\label{table:tuning_mrpc}
\end{table}

\begin{table}[t]
\centering
\scriptsize
\resizebox{1\linewidth}{!}{
\begin{tabular}{l|lllll}
\toprule
\diagbox{Baseline}{$\alpha$} &  1    & 5             & 10   & 20   \\
\midrule
\midrule
RoBERTa & \textbf{95.1} & 94.8 & 94.8 & 94.4 \\
\midrule
BART    & \textbf{96.2} & 96.1 & 95.6 & 95.3 \\
\bottomrule
\end{tabular}
}
\caption{Accuracy for Baseline+DefinitionEMB with various $\alpha$ on the SST validation set.}
\end{table}

\begin{table}[t]
\centering
\scriptsize
\resizebox{1\linewidth}{!}{
\begin{tabular}{l|lllll}
\toprule
\diagbox{Baseline}{$\alpha$}       & 3             & 5             & 10   & 20   \\
\midrule
\midrule
RoBERTa & 77.3          & \textbf{79.1} & 76.2 & 76.5 \\
\midrule
BART    & \textbf{85.9} & \textbf{85.9} & 84.8 & 84.1\\
\bottomrule
\end{tabular}
}
\caption{Accuracy for Baseline+DefinitionEMB with various $\alpha$ on the RTE validation set.}
\end{table}

\begin{table}[t]
\centering
\scriptsize
\resizebox{1\linewidth}{!}{
\begin{tabular}{l|lllll}
\toprule
\diagbox{Baseline}{$\alpha$}       & 3               & 5      & 7      & 10     \\
\midrule
\midrule
RoBERTa & \textbf{90.7} & 90.5 & 90.3 & 90.3 \\
\midrule
BART    & \textbf{91.8}   & 91.7   & 91.5   & 91.3  
\\
\bottomrule
\end{tabular}
}
\caption{Spearman’s rank correlation for Baseline+DefinitionEMB with various $\alpha$ on the STS validation set.}
\end{table}

\begin{table}[t]
\centering
\scriptsize
\resizebox{1\linewidth}{!}{
\begin{tabular}{l|lllll}
\toprule
\diagbox{Baseline}{$\alpha$}       & 3             & 7             & 10   & 30   \\
\midrule
\midrule
RoBERTa & \textbf{92.9} & \textbf{92.9} & 92.8 & 92.5 \\
\midrule
BART    & \textbf{94.8} & 94.7          & 94.5 & 94.4
\\
\bottomrule
\end{tabular}
}
\caption{Accuracy for Baseline+DefinitionEMB with various $\alpha$ on the QNLI validation set.}
\end{table}

\begin{table}[t]
\centering
\scriptsize
\resizebox{1\linewidth}{!}{
\begin{tabular}{l|lllll}
\toprule
\diagbox{Baseline}{$\alpha$}       & 3             & 5             & 10            & 30            \\
\midrule
\midrule
RoBERTa & \textbf{91.8} & 91.7 & \textbf{91.8} & \textbf{91.8} \\
\midrule
BART    & 92.4 & \textbf{92.6} & 92.5          & 92.2       \\
\bottomrule
\end{tabular}
}
\caption{Accuracy for Baseline+DefinitionEMB with various $\alpha$ on the QQP validation set.}
\end{table}

\begin{table}[t]
\centering
\resizebox{1\linewidth}{!}{
\begin{tabular}{l|lllll}
\toprule
\diagbox{Baseline}{$\alpha$}      & 3             & 5             & 10            & 30            \\
\midrule
\midrule
RoBERTa & \textbf{87.6} & 87.5 & 87.4 & 87.3 \\
\midrule
BART    & 89.7 & 89.6 & \textbf{89.8} & 89.6   \\
\bottomrule
\end{tabular}
}      
\caption{Accuracy for Baseline+DefinitionEMB with various $\alpha$ on the MNLI validation set.}
\end{table}

\begin{table}[t]
\centering
\scriptsize
\begin{tabular}{lccc}
\toprule
\multirow{2}{*}{$\alpha$} & \multicolumn{3}{c}{\textbf{ROUGE (F1)}}                    \\
                          & 1              & 2              & L              \\
\midrule
\midrule
10                        & 44.45          & \textbf{21.62} & 41.17          \\
30                        & 44.30          & 21.44          & 41.04          \\
50                        & 44.23          & 21.37          & 40.99          \\
100                       & \textbf{44.62} & 21.54          & \textbf{41.40} \\
\bottomrule
\end{tabular}
\caption{ROUGE scores for BART+DefinitionEMB with various $\alpha$ on the CNNDM validation set.}
\end{table}

\begin{table}[t]
\centering
\scriptsize
\begin{tabular}{lccc}
\toprule
\multirow{2}{*}{$\alpha$} & \multicolumn{3}{c}{\textbf{ROUGE (F1)}}                      \\
                          & 1              & 2              & L              \\
\midrule
\midrule
5                         & 44.02          & 20.66 & 34.95          \\
10                        & \textbf{44.22} & \textbf{20.95} & \textbf{35.24} \\
20                        & 43.81          & 20.55          & 34.78          \\
100                       & 42.46 & 19.18          & 33.62 \\
\bottomrule
\end{tabular}
\caption{ROUGE scores for BART+DefinitionEMB with various $\alpha$ on the XSUM validation set.}
\end{table}

\begin{table}[t]
\centering
\scriptsize
\begin{tabular}{lccc}
\toprule
\multirow{2}{*}{$\alpha$} & \multicolumn{3}{c}{\textbf{ROUGE (F1)}}                      \\
                          & 1              & 2              & L              \\
\midrule
\midrule
5                         & 50.63          & 32.19          & 38.81          \\
7                         & 50.85          & \textbf{32.44} & \textbf{39.10} \\
10                        & \textbf{51.08} & 32.17          & 38.97          \\
100                       & 50.04          & 31.43          & 38.27         \\
\bottomrule
\end{tabular}
\caption{ROUGE scores for BART+DefinitionEMB with various $\alpha$ on the Billsum validation set.}
\end{table}

\begin{table}[t]
\centering
\scriptsize
\begin{tabular}{lccc}
\toprule
\multirow{2}{*}{$\alpha$} & \multicolumn{3}{c}{\textbf{ROUGE (F1)}}                      \\
                          & 1              & 2              & L              \\
\midrule
\midrule
10                        & 43.62          & 18.53          & 37.43          \\
20                        & 43.93          & 18.84          & 37.74          \\
30                        & \textbf{44.22} & \textbf{19.12} & \textbf{38.03} \\
100                       & 42.96          & 17.76          & 36.73          \\
\bottomrule
\end{tabular}
\caption{ROUGE scores for BART+DefinitionEMB with various $\alpha$ on the Y-BIGPATENT validation set.}
\label{table:tuning_bigpatent}
\end{table}

\begin{table}[t]
\resizebox{1\columnwidth}{!}{
\begin{tabular}{lccccccc}
\toprule
\multirow{2}{*}{\textbf{Model}}                   & \multirow{2}{*}{\textbf{MRPC}} & \multirow{2}{*}{\textbf{SST}} & \multirow{2}{*}{\textbf{RTE}} & \multirow{2}{*}{\textbf{STS}} & \multirow{2}{*}{\textbf{QNLI}} & \multirow{2}{*}{\textbf{QQP}} & \multirow{2}{*}{\textbf{MNLI}}   \\
& & & & & & & \\
\midrule
\midrule
RoBERTa                &   5   &    1   &   5  &    3   &   3   &  3   &    3     \\
BART                   & 5    & 1     & 3   & 3     & 3    & 5   & 10              \\
\bottomrule
\end{tabular}
}
\caption{Tuned $\alpha$ for GLUE datasets.}
\label{table:tuned_alpha_glue}
\end{table}

\begin{table}[t]
\resizebox{1\columnwidth}{!}{
\begin{tabular}{lcccc}
\toprule
\multirow{2}{*}{\textbf{Model}}                   &  \multirow{2}{*}{\textbf{CNNDM}}  & \multirow{2}{*}{\textbf{Y-BIGPATENT}} & \multirow{2}{*}{\textbf{XSUM}}  & \multirow{2}{*}{\textbf{Billsum}}   \\
& & & & \\
\midrule
\midrule
BART                   & 100    & 30 & 10    & 7               \\
\bottomrule
\end{tabular}
}
\caption{Tuned $\alpha$ for text summarization datasets.}
\label{table:tuned_alpha_text_summarization}
\end{table}

\clearpage
\begin{figure*}[h]
    \centering
    \includegraphics[width=0.9\linewidth]{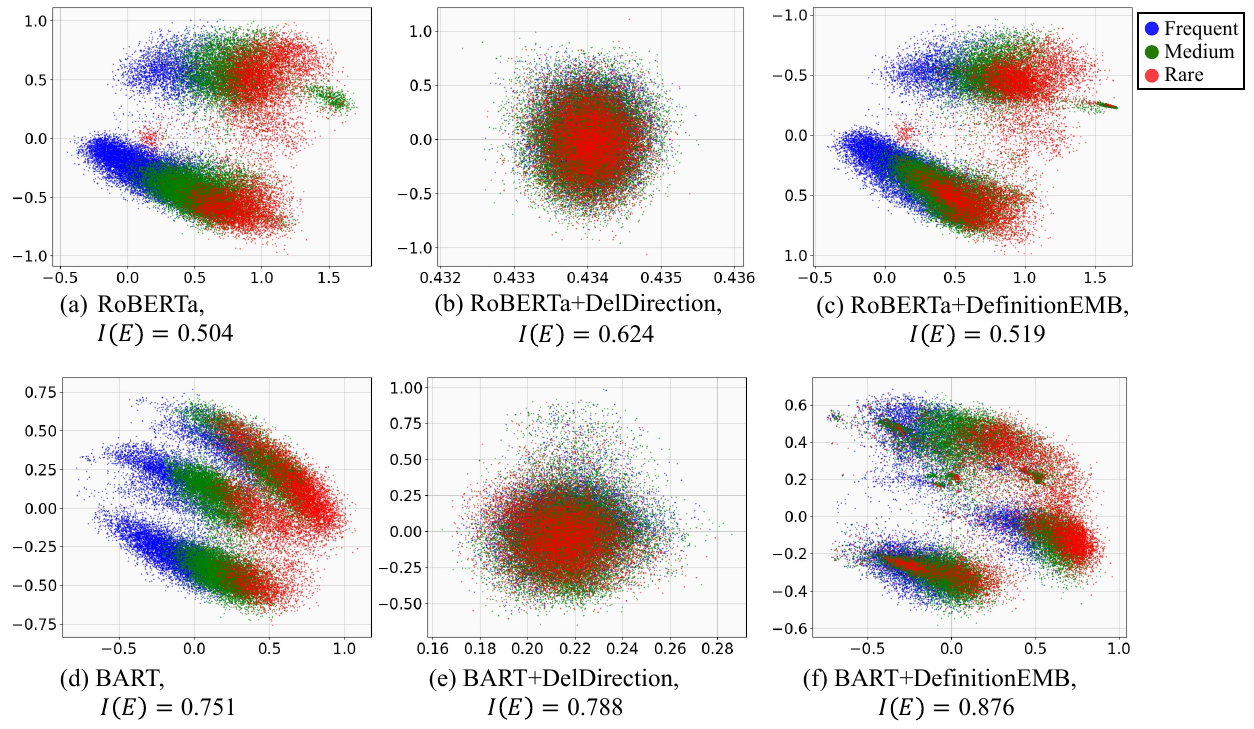}
\caption{Projected token embeddings of models before and after replacing $\mathbf{E}$ completely.
The frequent (30\%), medium (50\%), and rare (20\%) groups are determined based on the token index in $\mathcal{V}$.
}
\label{figure:projected_initial_embeddings}
\end{figure*}

\section{Additional Evaluation}
\subsection{Projected Initial Token Embeddings}
\label{appendix:projected_initial_token_embeddings}

Figure~\ref{figure:projected_initial_embeddings} shows the projected token embeddings of models before and after replacing $\mathbf{E}$ completely. 

\subsection{Differences in DefinitionEMB with Regard to Model Architecture}
\label{appendix:mse_between_embeddings}

\begin{figure*}[h]
\centering
\begin{subfigure}[t]{0.47\linewidth}
\begin{adjustbox}{width=1\columnwidth}
\begin{tikzpicture}
    \pgfplotsset{set layers}
    \begin{axis}[
            ybar,
            scale only axis,
            axis y line*=left,
            x post scale=1.25,
            y post scale=1,
            ylabel=\# of tokens in $\mathcal{V}$,
            ymin=0, ymax=11000,
            xlabel=MSE range,
            symbolic x coords={1,2,3,4,5,6,7,8,9,10,11,12,13,14,15,16,17,18,19},
            xtick=data,
            xticklabel style={rotate=90},
            xticklabels={[0\textbf{,}1),[1\textbf{,}2),[2\textbf{,}3),[3\textbf{,}4),[4\textbf{,}5),[5\textbf{,}6),[6\textbf{,}7),[7\textbf{,}8),[8\textbf{,}9),[9\textbf{,}10),[10\textbf{,}11),[11\textbf{,}12),[12\textbf{,}13),[13\textbf{,}14),[14\textbf{,}15),[15\textbf{,}16),[16\textbf{,}17),[17\textbf{,}18),[18\textbf{,}19)},
            bar width=5pt,
            font=\large,
        ]
        \addplot [blue, fill=blue, ybar, ybar legend] table[x=Position, y=count]{\MSERatioDistributionRoBERTA}; \label{mse_ratio_distribution_count_bar_roberta}
    \end{axis}
    \begin{axis}[
            legend pos= south east,
            scale only axis,
            axis y line*=right,
            x post scale=1.25,
            y post scale=1,
            ylabel=Cumulative ratio,
            axis x line=none,
            ymin=0, ymax=1.05,
            symbolic x coords={1,2,3,4,5,6,7,8,9,10,11,12,13,14,15,16,17,18,19},
            font=\large,
        ]
        \addplot+[sharp plot, mark = none,  red] table[x=Position,y=ratio]{\MSERatioDistributionRoBERTA};\label{mse_ratio_distribution_ratio_line_roberta}

    \end{axis}
\end{tikzpicture}
\end{adjustbox}
\end{subfigure}
\begin{subfigure}[t]{0.5\linewidth}
\begin{adjustbox}{width=1\columnwidth}
    \begin{tikzpicture}
    \begin{axis}[
        legend pos=outer north east,
        ybar stacked,
        ymin=0, ymax=16000,
        xtick=data,
        x post scale=1.2,
        y post scale=2,
        ylabel=\# of tokens in $\mathcal{V}$,
        xlabel=MSE range,
        reverse legend=true,
        xticklabels={[0\textbf{,}1),[1\textbf{,}3),[3\textbf{,}5),[5\textbf{,}8),[8\textbf{,})},
        xticklabel style={rotate=90},
        font=\huge,
    ]
    \addplot [fill=green!80] table [y=series1, meta=Label, x expr=\coordindex] {\MSEStackDistributionRoBERTA};
    \addlegendentry{Token index in [0, 10000)}
    \addplot [fill=blue!60] table [y=series2, meta=Label, x expr=\coordindex] {\MSEStackDistributionRoBERTA};
    \addlegendentry{Token index in [10000, 20000)}
    \addplot [fill=red!60,point meta=y] table [y=series3, meta=Label, x expr=\coordindex] {\MSEStackDistributionRoBERTA};
    \addlegendentry{Token index in [20000, 30000)}
    \addplot [fill=gray!60,point meta=y] table [y=series4, meta=Label, x expr=\coordindex] {\MSEStackDistributionRoBERTA};
    \addlegendentry{Token index in [30000, 40000)}
    \addplot [fill=orange!60,point meta=y] table [y=series5, meta=Label, x expr=\coordindex] {\MSEStackDistributionRoBERTA};
    \addlegendentry{Token index in [40000, 50000)}
    \end{axis}
    \end{tikzpicture}
\end{adjustbox}
\end{subfigure}
\caption{Number of tokens in $\mathcal{V}$ versus MSE estimated by DefinitionEMB based on RoBERTA model.}
\label{figure:bart_mse_desitribution_roberta}
\end{figure*}
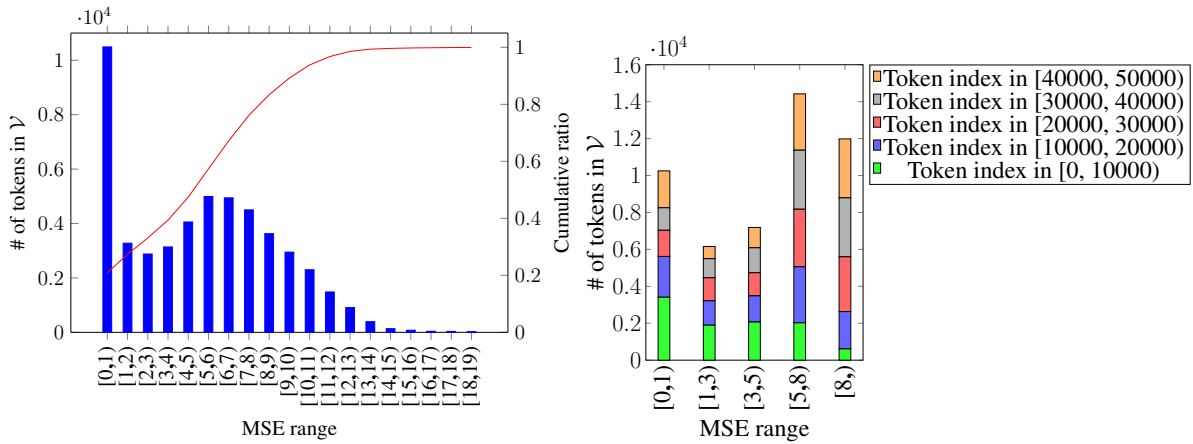

\begin{figure*}[h]
\centering
\begin{subfigure}[t]{0.47\linewidth}
\begin{adjustbox}{width=1\columnwidth}
\begin{tikzpicture}
    \pgfplotsset{set layers}
    \begin{axis}[
            ybar,
            scale only axis,
            axis y line*=left,
            x post scale=1.25,
            y post scale=1,
            ylabel=\# of tokens in $\mathcal{V}$,
            ymin=0, ymax=18000,
            xlabel=MSE range,
            symbolic x coords={1,2,3,4,5,6,7,8,9,10,11,12,13,14,15,16,17,18,19},
            xtick=data,
            xticklabel style={rotate=90},
            xticklabels={[0\textbf{,}1),[1\textbf{,}2),[2\textbf{,}3),[3\textbf{,}4),[4\textbf{,}5),[5\textbf{,}6),[6\textbf{,}7),[7\textbf{,}8),[8\textbf{,}9),[9\textbf{,}10),[10\textbf{,}11),[11\textbf{,}12),[12\textbf{,}13),[13\textbf{,}14),[14\textbf{,}15),[15\textbf{,}16),[16\textbf{,}17),[17\textbf{,}18),[18\textbf{,}19)},
            bar width=5pt,
            font=\large,
        ]
        \addplot [blue, fill=blue, ybar, ybar legend] table[x=Position, y=count]{\MSERatioDistribution}; \label{mse_ratio_distribution_count_bar}
    \end{axis}
    \begin{axis}[
            legend pos= south east,
            scale only axis,
            axis y line*=right,
            x post scale=1.25,
            y post scale=1,
            ylabel=Cumulative ratio,
            axis x line=none,
            ymin=0, ymax=1.05,
            symbolic x coords={1,2,3,4,5,6,7,8,9,10,11,12,13,14,15,16,17,18,19},
            font=\large,
        ]
        \addplot+[sharp plot, mark = none,  red] table[x=Position,y=ratio]{\MSERatioDistribution};\label{mse_ratio_distribution_ratio_line}

    \end{axis}
\end{tikzpicture}
\end{adjustbox}
\end{subfigure}
\begin{subfigure}[t]{0.5\linewidth}
\begin{adjustbox}{width=1\columnwidth}
    \begin{tikzpicture}
    \begin{axis}[
        legend pos=outer north east,
        ybar stacked,
        ymin=0, ymax=18000,
        xtick=data,
        x post scale=1.2,
        y post scale=2,
        ylabel=\# of tokens in $\mathcal{V}$,
        xlabel=MSE range,
        reverse legend=true,
        xticklabels={[0\textbf{,}1),[1\textbf{,}3),[3\textbf{,}5),[5\textbf{,}8),[8\textbf{,})},
        xticklabel style={rotate=90},
        font=\huge,
    ]
    \addplot [fill=green!80] table [y=series1, meta=Label, x expr=\coordindex] {\MSEStackDistribution};
    \addlegendentry{Token index in [0, 10000)}
    \addplot [fill=blue!60] table [y=series2, meta=Label, x expr=\coordindex] {\MSEStackDistribution};
    \addlegendentry{Token index in [10000, 20000)}
    \addplot [fill=red!60,point meta=y] table [y=series3, meta=Label, x expr=\coordindex] {\MSEStackDistribution};
    \addlegendentry{Token index in [20000, 30000)}
    \addplot [fill=gray!60,point meta=y] table [y=series4, meta=Label, x expr=\coordindex] {\MSEStackDistribution};
    \addlegendentry{Token index in [30000, 40000)}
    \addplot [fill=orange!60,point meta=y] table [y=series5, meta=Label, x expr=\coordindex] {\MSEStackDistribution};
    \addlegendentry{Token index in [40000, 50000)}
    \end{axis}
    \end{tikzpicture}
\end{adjustbox}
\end{subfigure}
\caption{Number of tokens in $\mathcal{V}$ versus MSE estimated by DefinitionEMB based on BART model.}
\label{figure:bart_mse_desitribution}
\end{figure*}
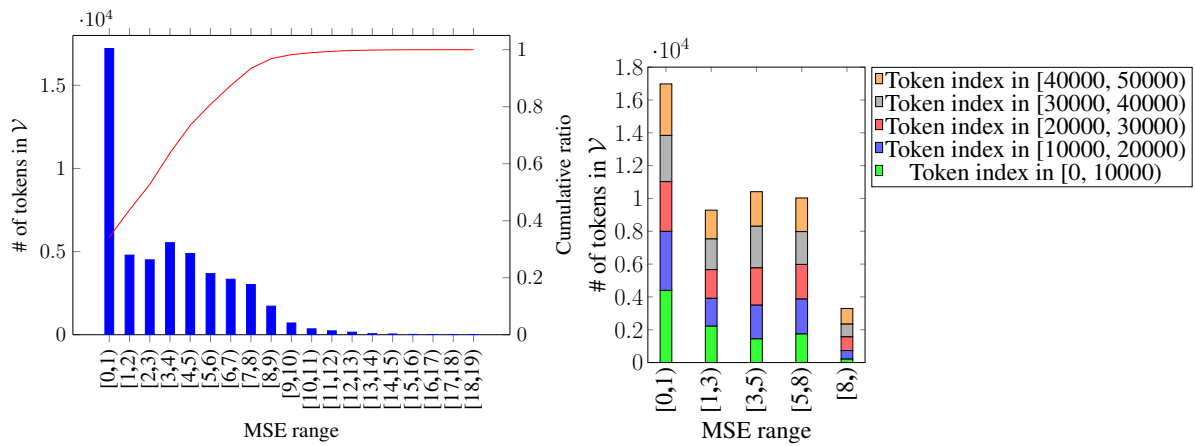

\begin{table}[h]
\centering
\begin{tabular}{lrr}
\toprule
\multirow{2}{*}{\textbf{Token}} & \multirow{2}{*}{\textbf{Index in $\mathcal{V}$}}   & \multirow{2}{*}{\textbf{MSE}} \\
& & \\
\midrule
\midrule
sys	& 43103 & 23.3829 \\
ĠNASL 	& 47179	& 23.2779 \\
resso 	& 27989	& 23.2549 \\
ĠFAQ 	& 39313	& 22.8146 \\
ĠpH 	& 39228	& 22.1278 \\
\midrule
ĠB 	& 163	& 0.0061 \\
ER    & 2076   & 0.0059 \\
s &  29	& 0.0055 \\
ING   & 1862   & 0.0051 \\
-  &  12 & 0.0040 \\
   \bottomrule
\end{tabular}
\caption{Top and bottom 5 tokens based on the degree of MSE estimated by DefinitionEMB based on RoBERTa model, listed in descending order of MSE.}
\label{table:example_mse_tokens_roberta}
\end{table}

\begin{table}[h]
\centering
\begin{tabular}{lrr}
\toprule
\multirow{2}{*}{\textbf{Token}} & \multirow{2}{*}{\textbf{Index in $\mathcal{V}$}}   & \multirow{2}{*}{\textbf{MSE}} \\
& & \\
\midrule
\midrule
ourke	& 18338 & 18.7777 \\
esson 	& 24711	& 17.6324 \\
aeus 	& 39174	& 17.3797 \\
wagen 	& 42099	& 16.8839 \\
auga 	& 24491	& 16.8624 \\
\midrule
ER 	& 2076	& 0.0020 \\
ES 	& 1723	& 0.0020 \\
ING 	& 1862	& 0.0020 \\
- 	&	12 & 0.0018 \\
S &  104	& 0.0018 \\
   \bottomrule
\end{tabular}
\caption{Top and bottom five tokens based on the degree of MSE estimated by DefinitionEMB based on BART model, listed in descending order of MSE.}
\label{table:example_mse_tokens}
\end{table}

\noindent \textbf{Different MSE Distributions.} For tokens in PLM's vocabulary $\mathcal{V}$, we analyze the mean squared error (MSE) between their pre-trained and definition embeddings. 
Figure~\ref{figure:bart_mse_desitribution_roberta} presents the results for DefinitionEMB on the RoBERTa model. The left subfigure illustrates that around 20\% of the tokens have an MSE of less than 1, while less than 20\% tokens have an MSE larger than 8. 
The right subfigure shows that the distribution of token index is almost uniform across the MSE, indicating that the pre-trained embedding of high-frequency tokens may contain semantically unrelated information, while the pre-trained embedding of low-frequency tokens may contain semantically related information even with limited pre-training steps. In addition, more tokens fall in the MSE range of [5, 8) than in the range of [0, 1), which indicates a significant difference between pre-trained embeddings and definition embeddings.
Figure~\ref{figure:bart_mse_desitribution} presents the results for DefinitionEMB on the BART model. The left subfigure illustrates that around 40\% of the tokens have an MSE of less than 1, while less than 10\% tokens have an MSE larger than 8. The right subfigure also shows that the distribution of token index is almost uniform across the MSE. However, most of the tokens fall in the MSE range [0, 1), indicating less difference between pre-trained embeddings and definition embeddings than in the case of RoBERTa. Tables~\ref{table:example_mse_tokens_roberta} and~\ref{table:example_mse_tokens} list examples of tokens with the corresponding MSE. The top five tokens with the highest MSE can be used as named entities.\newline
\newline
\noindent \textbf{Reasons.} The MSE results align with the original representation distribution in PLMs, where RoBERTa has fewer embedding parameters and exhibits lower isotropy than BART (Table~\ref{table:isotropy_results}). Because DefinitionEMB considers the pre-trained embeddings as gold embeddings, its constructed embeddings for RoBERTa naturally exhibit lower isotropy and higher MSE than those for BART. Additionally, the different masking strategies for encoder-only and encoder-decoder architectures may also lead to the different MSE and isotropy distributions, as~\citet{10.5555/3455716.3455856} demonstrate that the T5-style masking mechanism is more effective than the BERT-style masking for the encoder-decoder model during pre-training.

\subsection{Word Similarity Task Using Cosine Similarity}
\label{appendix:word_similarity}
\begin{table}[h]
\centering
\resizebox{1\columnwidth}{!}{
\begin{tabular}{lccccc}
\toprule
\multirow{2}{*}{\textbf{Model}} & \multicolumn{5}{c}{\textcolor{black}{\textbf{Spearman Score} $\uparrow$}}\\
\cline{2-6}
& \textbf{RG65}  & \textbf{RW}   & \textbf{SimLex}  & \textbf{SimVerb}  & \textbf{Ave} \\
\midrule
\midrule
RoBERTa & 25.55 & 22.33 & \textbf{18.04} & \textbf{10.78} & 19.18 \\
+ DefinitionEMB & \textbf{29.56} & \textbf{22.82} & 17.48 & 10.58 & \textbf{20.11} \\
\midrule
BART       & \textbf{24.35}    & 23.20     &      \textbf{22.00}      & \textbf{12.73}  & \textbf{20.57} \\
+ DefinitionEMB & 22.93 & \textbf{23.40}   &      21.04            & 12.09 &  19.87\\
\bottomrule
\end{tabular}
}
\caption{Experimental results on the word similarity task with cosine similarity. DefinitionEMB replaces $\mathbf{E}$ completely.}
\label{table:appendix_word_similarity}
\end{table}

\noindent Table~\ref{table:appendix_word_similarity} shows the results for word similarity tasks, where the similarity between word embeddings is calculated using cosine similarity.
Comparing Table~\ref{table:appendix_word_similarity} with Table~\ref{table:word_similarity}, we observe that when using the dot product, DefinitionEMB achieves a higher Spearman score than BART. However, when using cosine similarity, the opposite result is observed. It indicates that the constructed embeddings for BART prioritize distance over angle aspects from the pre-trained embeddings.

\subsection{Isotropy on GLUE Task}
\label{appendix:isotropy_on_GLUE}

\begin{table*}[t]
\resizebox{1\linewidth}{!}{
\begin{tabular}{lllllllllllllll}
\toprule
\multirow{2}{*}{\textbf{Model}} & \multicolumn{2}{c}{\textbf{SST}} & \multicolumn{2}{c}{\textbf{MRPC}} & \multicolumn{2}{c}{\textbf{STS}} & \multicolumn{2}{c}{\textbf{QQP}} & \multicolumn{2}{c}{\textbf{MNLI}} & \multicolumn{2}{c}{\textbf{QNLI}} & \multicolumn{2}{c}{\textbf{RTE}} \\
                          & Before          & After       & Before          & After          & Before           & After          & Before          & After          & Before           & After          & Before           & After          & Before          & After          \\
                                \midrule
                                \midrule
RoBERTa & 0.504 & 0.533 & - & 0.505 & - & 0.506 & - & 0.542 & - & 0.544 & - & 0.509 & - & 0.505 \\
+DelDirection & 0.624 & 0.627 & - & 0.624 & - & 0.625 & - & 0.642 & - & 0.639 & - & 0.629 & - & 0.625\\
+DefinitionEMB & 0.528 & 0.536& 0.529& 0.529 &0.529 & 0.530 & 0.530 & 0.550 & 0.530& 0.554&0.539 & 0.541 & 0.529 & 0.533 \\
\midrule
  BART                          &  0.751              &    0.751            &  -  &     0.751           &       -           &      0.751          &     -            &      0.752          &        -          &          0.751      &        -          &      0.751         &   -              &  0.751  \\
  +DelDirection                          &   0.788        &      0.788          &    -     &     0.788           &         -         &        0.788        &       -          &    0.800            &       -           &        0.805        &        -          &       0.794         &        -         &   0.788 \\
  +DefinitionEMB                          &  0.766     &    0.766            &       0.767      &   0.767             &       0.769           &     0.769           &     0.769            &     0.777           &      0.770            &       0.771         &        0.753          &         0.753       &           0.753      &   0.753 \\
                                \bottomrule
\end{tabular}
}
\caption{$I(\mathbf{E})$ for models before and after fine-tuning on the GLUE task.}
\label{table:isotropy_vtrain_glue}
\end{table*}

Table~\ref{table:isotropy_vtrain_glue} presents $I(\mathbf{E})$ for models before and after fine-tuning on the GLUE task. Because $I(\mathbf{E})$ of PLMs and the DelDirection model before fine-tuning does not depend on $\mathcal{V}_{[task]}$, it is reported only once in the table. 
For the MRPC, STS, and RTE datasets, $I(\mathbf{E})$ shows a minimal difference between before and after fine-tuning \textcolor{black}{models}, likely due to the limited number of fine-tuning steps on these datasets.
The token embedding distribution in BART appears to be more stable than RoBERTa on the SST, QQP, and MNLI datasets.
Using DelDirection for RoBERTa and BART achieves the highest $I(\mathbf{E})$ across all datasets; however, it also results in the lowest accuracy and Pearson/Spearman’s rank correlation in most cases, as shown in Table~\ref{table:GLUE_results}.
This supports our assumption that DelDirection model focuses on the distribution of embeddings at the expense of semantic information.

\subsection{Ablation Study for Replacing Tokens}
\label{appendix:ablation_mrpc}

\begin{table}[h]
\centering
\resizebox{1\columnwidth}{!}{
\begin{tabular}{lcc}
\toprule
\textbf{Replaced} &   \textbf{Y-BIGPATENT} & \textbf{Billsum} \\
\textbf{tokens}    & ($X=23,000$) & ($X=41,000$) \\
\midrule
\midrule
 Appearing &   \textbf{44.16} / \textbf{19.06} / \textbf{38.01}              &       50.96 / \textbf{32.64} / \textbf{39.28}                               \\
Both   &  44.00 / 18.90 / 37.84                  &  \textbf{51.23} / 32.44 / 39.20          \\
\bottomrule
\end{tabular}
}
\caption{Replacing appearing tokens only vs. replacing both appearing and non-appearing tokens for BART \textcolor{black}{on the Y-BIGPATENT and Billsum test sets}.
}
\label{table:replace_versus_bigpatent_billsum}
\end{table}

\begin{table}[h]
\centering
\resizebox{0.8\columnwidth}{!}{
\begin{tabular}{llc}
\toprule
\multirow{2}{*}{\textbf{Model}}     & \multirow{2}{*}{\textbf{Replaced tokens}}    & \multirow{2}{*}{\textbf{MRPC}} \\
&&\\
\midrule
\midrule
RoBERTa& Appearing & \textbf{87.7}\\
& Both & 87.3 \\
\midrule
BART & Appearing &    \textbf{88.3}                                        \\
& Both &   88.1                    \\
\bottomrule
\end{tabular}
}
\caption{Replacing appearing tokens only vs. replacing both appearing and non-appearing tokens for RoBERTa and BART \textcolor{black}{on the MRPC test set with $X=24,900$}.}
\label{table:replace_versus_mrpc}
\end{table}

We conducted an ablation study to analyze the effectiveness of replacing only appearing tokens instead of all tokens. The index range of replaced tokens is denoted as $[X, |\mathcal{V}|]$, and the number of tokens appearing in $[X, |\mathcal{V}|]$ satisfies $\min(\alpha \% * |\mathcal{V}|, |\mathcal{V}_{[task]}|)$, as required in Section~\ref{sec:replacing_strategy}.

Results for BART+DefinitionEMB are reported in \textcolor{black}{Table~\ref{table:replace_versus_bigpatent_billsum}}. 
When replacing only appearing tokens, the model achieves higher ROUGE scores than when replacing all tokens. Specifically, ROUGEL-F1 improved by 0.17 and 0.08 for Y-BIGPATENT and Billsum, respectively. This difference may be caused by the varying token frequencies in the training sets.
\textcolor{black}{Table~\ref{table:replace_versus_mrpc} shows the ablation study with respect to the replacing strategies for Baseline+DefinitionEMB on the MRPC test set.
When replacing only appearing tokens, both RoBERTa and BART yield higher accuracy scores.
}

\clearpage

\section{Insights into Rare Tokens in Fine-tuned Models}
This section analyzes whether fine-tuned models can distinguish rare tokens and effectively extract their semantics during text generation.

\subsection{Probing Numeric-related Semantics.}
\label{appendix:details_on_probing}

\begin{table}[h]
\resizebox{1\linewidth}{!}{
\begin{tabular}{lrrr}
\toprule
\multirow{2}{*}{\textbf{Label}}            & \multirow{2}{*}{\textbf{\# of training}} & \multirow{2}{*}{\textbf{\# of validation}} & \multirow{2}{*}{\textbf{\# of test}} \\
& & & \\
\midrule
\midrule
Numeric \textless 1000    & 836                     & 93                        & 72                  \\
Numeric \textgreater 1000 & 878                     & 98                        & 6                   \\
Others                    & 900                     & 100                       & 78     \\
\midrule
All & 2614 & 291 & 156 \\
\bottomrule
\end{tabular}
}
\caption{Numeric dataset statistics.}
\label{table:numeric_dataset}
\end{table}

\noindent To investigate whether fine-tuned models can distinguish between rare numeric and non-numeric tokens and effectively extract rare numeric tokens during text generation, we conduct a probing test. Specifically, we feed tokens into a CNNDM fine-tuned model and train an additional linear classifier on the model’s final hidden layer to predict three target labels: ``numeric < 1,000'', ``numeric > 1,000'', and ``others''. The details of the experimental settings are as follows. \newline
\newline
\noindent \textbf{Dataset.} The dataset statistics for this probing test are shown in Table~\ref{table:numeric_dataset}. We first introduce our test set.
As mentioned in Appendix~\ref{appendix:hyperparameters}, we added definitions for 1,252 numbers in $\mathcal{V}$ by translating numbers into their corresponding words, such as ``2'' to ``two''. 
Out of these 1,252 numbers, 78 numeric tokens start with ``Ġ''and appear in the CNNDM dataset with an index greater than 40,000 in the vocabulary $\mathcal{V}$.\footnote{Note that tokens starting with or without ``Ġ'' are mapped into different embedding regions.
Classifying both of them into the same label would place an unnecessary burden on the model during probing, so we only use those starting with``Ġ'' here.}
We also randomly select 78 non-numeric tokens from $\mathcal{V}$, ensuring they start with ``Ġ'' and  belong to the rare group within the CNNDM dataset. These 156 tokens make up our test sets. 

For the training and validation sets, we first randomly select 1,000 non-numeric words from the Wiktionary dataset, ensuring that a whitespace is added at the beginning of each word. This ensures that after tokenization, the first subword token will start with ``Ġ''. Similarly, we select numbers in the range [0, 1999] as our numeric words and added whitespace before them. These words were then divided into training and validation sets in a 9:1 ratio.
We further filter the data to ensure no overlap between the training/validation and the test data.\newline
\begin{table}[ht]
\centering
\resizebox{0.8\linewidth}{!}{
\begin{tabular}{lr}
\toprule
\multirow{2}{*}{\textbf{Hyperparameters}} & \multirow{2}{*}{\textbf{Value}}\\
& \\
\midrule
\midrule
\# of updates & 2,810  \\
\# of warm-up updates & 169 \\
\# of epochs & 10 \\
Batch size (sentences) & 512 \\
Learning rate & 7e-4 \\
Learning rate schedule & Polynomial decay \\
Optimizer & Adam \\
\bottomrule
\end{tabular}
}
\caption{Hyperparameters used for probing.}
\label{table:Hyperparameters_probing}
\end{table}

\noindent \textbf{Model Settings.} For both the training and validation sets, each tokenized word is fed into the encoder and decoder of the CNNDM fine-tuned model. Similarly, for the test set, each numeric or non-numeric token is processed by the encoder and decoder. Classification is performed using hidden states at the final token position in the last layer of the decoder to predict the three target labels: ``numeric < 1,000'', ``numeric > 1,000'', and ``others''. During training, the fine-tuned parameters are frozen, and an additional feed-forward neural network is trained as the classifier. This classifier consists of two layers and is open-sourced.\footnote{\url{https://github.com/facebookresearch/fairseq/blob/main/fairseq/models/bart/model.py\#L294}} Table~\ref{table:Hyperparameters_probing} lists the training parameters used for classification across the three labels.

\newpage
\subsection{Sample of Summarization}
\label{appendix:sample_of_summarization}

\begin{table*}[]
\resizebox{\textwidth}{!}{
\begin{tabular}{p{2.5cm}|p{15.5cm}}
\toprule
Source         & (CNN)For superhero fans, the cup runneth over. Most of us know the members of the Avengers by now: Iron Man, Captain America, Hulk and the rest, and the fact that a few more like Quicksilver are joining the cast \textbf{\textcolor{blue}{in the "Avengers: Age of Ultron" sequel}}. But there was one character who remained a mystery: \textbf{\textcolor{blue}{the Vision, to be played by Paul Bettany.}} Thus far, we've only seen his eyes in a trailer. With less than a month to go before the movie hits theaters, \textbf{\textcolor{blue}{Marvel Studios put all the speculation to rest with a poster featuring Bettany as the heroic android}}, who was a member of the superhero group for many years in the comics. Meanwhile, as many Marvel fans know, \textbf{\textcolor{blue}{Thursday was the eve of the new Netflix series "Daredevil," and after a photoshopped first look at Charlie Cox's iconic red Daredevil suit went out, Marvel put out a video of the real one.}} Not to be outdone, \textbf{\textcolor{blue}{director Bryan Singer announced a new character for next year's sequel "X-Men: Apocalypse," by telling Empire magazine that Ben Hardy would be playing the role of the winged mutant Angel.}} He even had a photo to share. And Thursday's new super images weren't quite done, because the questions over how \textbf{\textcolor{blue}{Jamie Bell's rocky character The Thing in the rebooted "Fantastic Four" movie (out August 7) might look were also finally answered}}. And he looks ... pretty much like The Thing we already knew (but reportedly, CGI this time). Within 24 hours, we got yet another indication that the superhero trend isn't going anywhere anytime soon (and we didn't even talk about the new photo of Ryan Reynolds' "Deadpool"). \\
\midrule
Reference         & \textcolor{red}{\underline{Marvel}} Studios releases first looks at Paul Bettany as the Vision in "Avengers: Age of\textcolor{red}{\underline{ Ultron}}" and Charlie Cox in full "D\textcolor{red}{\underline{aredevil}}" costume . Jamie Bell's character of The Thing was also unveiled for 20th Century Fox's Marvel-based reboot of "Fantastic Four" Bryan Singer unveiled the first look at "X-Men:\textcolor{red}{\underline{ Apocalypse}}" Angel played by Ben Hardy .  \\
\midrule
BART           & Paul Bettany will play the Vision in the "Avengers: Age of\textcolor{red}{\underline{ Ultron}}" sequel . The actor has been playing the\textcolor{red}{\underline{ android}} for many years in the comics . The "Fantastic For" reboot's" The Thing" looks pretty much like The Thing we already knew .\\
\midrule
+DelDirection   & Paul Bettany's character in "Avengers: Age of\textcolor{red}{\underline{ Ultron}}" is finally revealed . The actor has been playing the Vision in the comics for many years . The "Fantastic Fou" reboot's" The Thing" looks pretty much like The Thing we already knew (but CGI)\\
\midrule
+DefinitionEMB  & Paul Bettany will play the Vision in the "Avengers: Age of\textcolor{red}{\underline{ Ultron}}" sequel . Marvel Studios also announced a new character for "X-Men:\textcolor{red}{\underline{ Apocalypse}}" Ben Hardy will play the winged\textcolor{red}{\underline{ mutant}} Angel in "X-Men:\textcolor{red}{\underline{ Apocalypse}}," director Bryan Singer said .\\
\bottomrule
\end{tabular}
}
\caption{\textcolor{black}{Sample summarization of CNNDM test set. 
\textcolor{blue}{\textbf{Bold}} in source indicates the reference-related text. 
\textcolor{red}{\underline{Underline}} in reference and model outputs indicates the rare token with index larger than 40,000 in PLM's vocabulary $\mathcal{V}$.}}
\label{table:sample_cnndm_index40000}
\end{table*}

\begin{table*}[]
\resizebox{\textwidth}{!}{
\begin{tabular}{p{2.5cm}|p{15.5cm}}
\toprule
Source         & (CNN)It would have made Thomas Jefferson proud. Established on the birthday of the American founding father, Liberland -- the world's newest micronation -- is founded on a firm belief in liberty and noninterference from the powers-that-be. A tiny, 7 square-kilometer parcel of land, marked on maps as Gornja Siga, its territory abuts the Danube \textbf{\textcolor{blue}{on the border between Serbia and Croatia.}} The victim of a border dispute between Serbia and Croatia, it is claimed by neither side -- effectively a no-man's land. No one lives on this patch of land, which is heavily forested and contains only badly-maintained access roads and a run-down house, abandoned for decades. \textbf{\textcolor{blue}{This is where Euroskeptic Czech politician Vit Jedlicka stepped in.}} \textbf{\textcolor{blue}{On April 13 he planted his newly-designed yellow and black flag in the territory, declaring the area the Free Republic of Liberland}} -- a tiny sliver of a country, bigger only than the Vatican and Monaco. \textbf{\textcolor{blue}{He tells CNN that the country will be formally founded on May 1}} and is inviting, through the media, the world's heads of state to attend a formal ceremony marking the presumptive nation's birth. He says that he will also invite 7,500 of the 300,000 applicants that applied to become citizens of Liberland to the ceremony, where he will grant them citizenship. "I will grant citizenship if they can make it to the party," he told CNN by phone. "It's short notice but a good challenge, and also for  the presidents (and other heads of state) if they can make it to the founding of our country." Jedlicka, an active member of the Czech Republic's Party of Free Citizens, opposes excessive government interference. He says his attempts to enact change in his home country led him to the political experiment that is Liberland. "I would describe it as a global revolution. It's just the beginning," he tells CNN via Skype. Founded on staunchly libertarian principles -- its motto is "To live and let live" -- its website describes its system of governance as being a "constitutional republic with elements of direct democracy." It will use a form of cryptocurrency -- similar to Bitcoin -- as its national currency, bypassing the need for a central bank and will, according to its constitution, keep government's noses out of everything possible, from the banks to prostitution. "Liberland prides itself on personal and economic freedom of its people, which is guaranteed by the Constitution, which significantly limits the power of politicians so they could not interfere too much in the freedoms of the Liberland nation," the world's newest constitutional document states. Financial regulation will be minimal, if at all present. \textbf{\textcolor{blue}{Jedlicka says almost 300,000 applications for citizenship have been received}}, about 20 of which have been accepted. "Thousands of Americans, Swiss people. Also a lot of Arabic peoples who feel oppressed by the regimes there." He envisions, ultimately, a community of around 35,000 Liberlanders, not all of whom will be full-time residents. He says he expects trouble from his neighbors, whose land he has effectively annexed. "From Serbia, Croatia, we expect some trouble but we expect international laws will applied and any movement against us would be an attack on a sovereign nation, and we will offer nothing but passive resistance. For now, (though) we will make roads, docks." For its part, the Serbian Ministry of Foreign Affairs released a statement to CNN that stated that Liberland would not theoretically impinge upon its border, which is delineated by the Danube, but "the Ministry also considers this a frivolous act which needs no further comment." Croatia's counterpart was similarly dismissive. "Virtual quips, however interesting they occasionally sound, remain what they are -- virtual quips, and for them we have no official comment." \\
\midrule
Reference         & Vit Jedlicka, the first president of Liberland, tells CNN that the country will be formally founded on May\textcolor{red}{\underline{ 1}} . On April\textcolor{red}{\underline{ 13}}, Jedlicka declared an area between Croatia and Serbia "the Free Republic of Liberland" Jedlicka says that almost\textcolor{red}{\underline{ 300}},\textcolor{red}{\underline{000}} applications for citizenship have so far been received .\\
\midrule
BART           & Liberland is a tiny,\textcolor{red}{\underline{ 7}} square-kilometer parcel of land, marked on maps as Gornja Siga, which abuts the Danube on the border between Serbia and Croatia . The land is claimed by neither Serbia or Croatia, and is effectively a no-man's land . Czech politician Vit Jedlicka planted his newly-designed yellow and black flag in the territory, declaring the area the Free Republic of Liberland . He says he expects trouble from his neighbors, whose land he has effectively annexed .\\
\midrule
+DelDirection   & Czech politician Vit Jedlicka has declared the Free Republic of Liberland, a tiny sliver of land on the border of Serbia and Croatia . He says he expects trouble from his neighbors, whose land he has effectively annexed . "I would describe it as a global revolution. It's just the beginning," he tells CNN .\\
\midrule
+DefinitionEMB  & Czech politician Vit Jedlicka has declared the Free Republic of Liberland, a tiny sliver of land on the Serbian-Croatian border . He says he will grant citizenship to\textcolor{red}{\underline{ 7}},\textcolor{red}{\underline{500}} of the\textcolor{red}{\underline{ 300}},\textcolor{red}{\underline{000}} applicants who applied to become citizens of Liberland . "I would describe it as a global revolution. It's just the beginning," says Jedlicka via Skype .\\
\bottomrule
\end{tabular}
}
\caption{\textcolor{black}{Sample summarization of CNNDM test set. 
\textcolor{blue}{\textbf{Bold}} in source indicates the reference-related text. 
\textcolor{red}{\underline{Underline}} in reference and model outputs indicates the numeric in PLM's vocabulary $\mathcal{V}$.}}
\label{table:sample_cnndm_number}
\end{table*}

Tables~\ref{table:sample_cnndm_index40000} and~\ref{table:sample_cnndm_number} show sample summarizations of CNNDM test set. These examples show that using DefinitionEMB helps the PLM to understand and generate rare tokens.

\subsection{Isotropy of Contextual Token Representations.}
Figure~\ref{fig:DefinitionEMB_layer_cnndm_after} depicts the projection of contextual token representations. That is the specific decoder layer hidden states of the token in a given context~\citep{cai2021isotropy}. Although BART + DelDirection (Figure~\ref{fig:embedding_dynamics} (g)) exhibits a totally different token embedding distribution from BART (Figure~\ref{fig:embedding_dynamics} (b)), it yields a similar contextual representations as BART. Specifically, Figures~\ref{fig:DefinitionEMB_layer_cnndm_after} (a) and (b) show that high-frequency tokens are closely grouped together based on their frequency, while low-frequency tokens are spread out in specific directions. However, when using DefinitionEMB, we observe a more concentrated distribution than BART. Its projection resembles concentric ellipses, where tokens with similar frequencies are placed in the same ellipse.
This indicates that the last hidden states in DefinitionEMB involve less frequency-related information, allowing tokens of different frequencies to mix together. This finding aligns with our experimental results in Figure~\ref{figure:probing_results}, where using DefinitionEMB enables the model to encode and extract more semantic information than both BART and DelDirection. Combing all previous experiments, we conclude that using DefinitionEMB helps the PLMs to distinguish rare tokens and effectively extract rare tokens' semantics for text generation.

\begin{figure*}
\centering
    \includegraphics[width=1\linewidth]{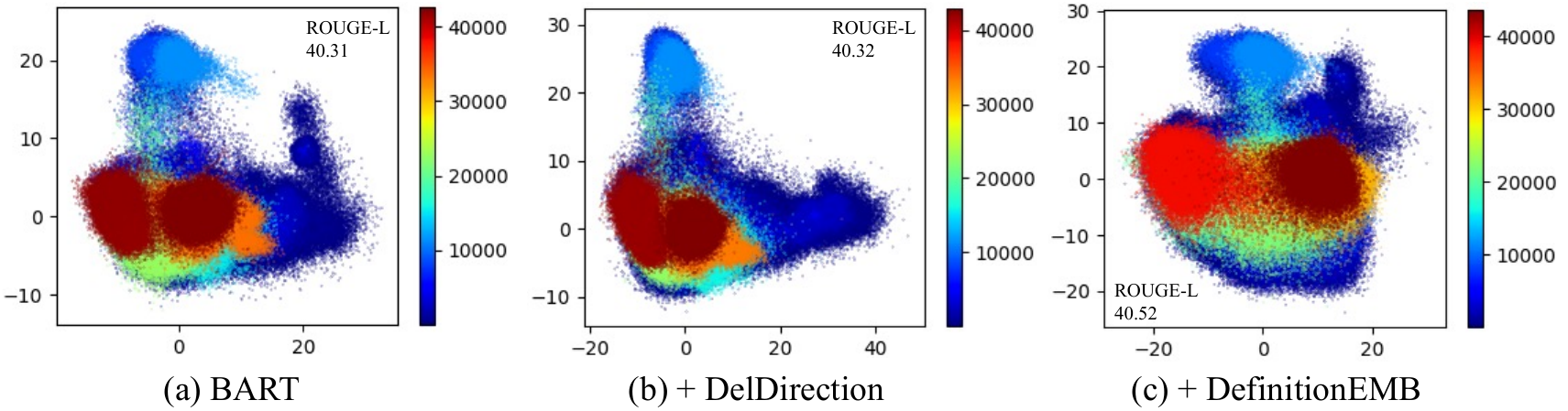}
    \caption{Projected contextual embedding in the 12th (final) decoder layer of \textcolor{black}{BART-related} models after fine-tuning on the CNNDM dataset. Colors indicate the token frequencies in corresponding test set.
    }
    \label{fig:DefinitionEMB_layer_cnndm_after}
\end{figure*}

\end{document}